\documentclass[journal]{IEEEtran}
\usepackage{amsmath,amssymb,amsfonts}
\usepackage{graphicx}
\usepackage{booktabs}
\usepackage{array}
\usepackage{float}
\usepackage{algorithm}
\usepackage{algorithmic}

\usepackage[hidelinks]{hyperref}

\begin{document}
\title{SCOPE: Certified Theorem Proving with a Language Model as the Policy Planner}
\pagestyle{plain}
\author{Hanchao Zhou and Jialei Li\thanks{Jialei Li (e-mail: gayly0414@163.com). Corresponding author: Hanchao Zhou (e-mail: zhouhanchao6@gmail.com).}}
\maketitle
\thispagestyle{plain}
\begin{abstract}In proof assistants such as Lean, a generated proof must pass machine compilation, so evaluation needs no human scoring. Direct generation fails on multi-step numeric propositions: a proof is valid only if every load-bearing integer is correct; treating per-integer errors as independent, the pass rate is bounded by the k-th power of the per-integer accuracy. Controlled corruption across 2,617 reference proofs supports the power law and locates its deviations at non-load-bearing positions. SCOPE (State-Conditioned Operator Planning and Execution) enforces the natural division of labor: the model plans over an operator vocabulary, a symbolic execution engine executes the numerics, and a compiler renders the proof. On a 218-problem suite it certifies 191/218 (87.6\%) with a 135M backbone; the 7B DeepSeek-Prover-V1.5-RL certifies 18/218 at 37.5 times the wall-clock, and DeepSeek-Prover-V2-7B certifies zero on a bidirectional dual suite. Multi-step thinking costs 6.12 decision actions per problem and produces no thinking text. Replacing the lagged engine state with the current one lifts the pass rate from 117/218 to 191/218, while up-weighting the chain-end loss lowers it. On the Lean-Workbook library, 2,132 of the 3,536 gradeable problems certify (60.29\%) under the full pipeline; on the extension families, the model's own chains certify 45.10\% of the 623 renderable rows, matching the family-tree chains (44.07\%), with no significant regression on the main suite. All certification readings come from a version-frozen review with independent rechecks and reverse verification. Within this domain, restricting free generation and keeping decision-time information visible is a more direct route than enlarging the model.\end{abstract}
\section{Introduction}
Automatic generation of formal proofs has long been a hard problem at the intersection of artificial intelligence and formal verification. Proof assistants such as Lean \cite{lean} bring propositions, proofs, and the checking process into one machine-verifiable framework: a candidate proof is accepted only if the compiler confirms that every step is legal and that no unproven assumption is used. This property makes theorem proving a suitable testbed for studying the reasoning ability of language models: evaluation involves no human scoring, and surface fluency earns no credit.
In recent years, language models ranging from hundreds of millions to tens of billions of parameters have made a series of advances in this direction, yet certified pass rates on propositions that require multi-step numeric derivation remain low. Failed samples show a consistent pattern: the proof is structurally conventional (it uses standard tactics such as induction, rewriting, and normalization) but breaks at concrete numeric values and ends with an unsolved goal (sorry in Lean). The pattern suggests that the bottleneck may lie not in the choice of proof steps but in generating the numeric content one item at a time. The division of labor can be stated quantitatively: the validity of a long proof is the conjunction of its load-bearing numeric content. If the model writes each load-bearing element correctly with probability p and a proof contains k such elements, the pass rate of direct generation is bounded above by $p^{k}$. When k grows with problem difficulty, any fixed p below one drives the pass rate down exponentially. Our experiments support this analysis and fold the countermeasure into the same architectural frame: if the analysis holds, the rational architectural response is to narrow the model's role to policy planning free of numeric generation, hand numeric computation to a symbolic tool, and hand proof text to template compilation.
In human mathematical practice, the core competence of solving lies to a large extent in perceiving and choosing tools: judging which theorem, which lemma, which transformation, or which algorithm applies to the proposition at hand. Carrying out the computation is comparatively mechanical and has always been done precisely by external procedures. We therefore test a division-of-labor hypothesis: let the language model supply the connection between problem and tool, and let a symbolic computing component execute the numerics; then examine whether this division supports long-chain, multi-step, exact reasoning.
Specifically, on three families of constructively generated discrete-mathematics problems (induction, inequalities, and number theory), we build a complete system called SCOPE (State-Conditioned Operator Planning and Execution). The backbone is the 135M-parameter SmolLM2 \cite{smollm2}, fitted with an operator-selection head and an argument-prediction head that produce a solution chain step by step inside a vocabulary of 194 operators; the numerics of each step are computed by a symbolic execution engine that returns the state; the proof text is generated by a compiler from the chain and the engine's results, and is finally reviewed by the Lean compiler. On this architecture we run a set of experiments that isolate one variable at a time, plus an external-model comparison. The main results follow.
When complete proof text is generated directly, token-level accuracy and certified pass rate diverge sharply: the model's token accuracy on 48 held-out rows from the training distribution is 0.9637, yet certified passes on the 218-problem suite number 0/218; a same-distribution control set (20 problems) yields one pass, consistent with the power-law prediction of about 0.031 computed from the number of content elements per proof (about 14.9). The 7B DeepSeek-Prover-V1.5-RL \cite{dsp15} passes 18/218 on the same suite, with 177 outputs terminating in sorry, the same pattern; DeepSeek-Prover-V2-7B certifies zero on both the main suite and a stratified public-library suite. Once proof text is produced by the compiler instead, the certified pass rate on the same suite rises from 1/218 (the slotting baseline, Experiment 2) to 117/218; feeding the chains of the pre-existing model to the compiler already yields 121/218 (Section VII). The compiler reproduces the reference proofs byte for byte on all 2,399 corpus rows and certifies 218/218 when fed the reference chains of the test suite, so this stage contributes no precision loss.
Analysis of the remaining errors further localizes the failure to the final step of the chain: all 101 failed problems make their first error at the last step. The residual class read off the final step stands in a fully determined correspondence with the engine's residual value on every sum-of-squares row (799/799 on the corpus and 181/181 on the test suite), but the decision frame carries a state that lags one step behind. A $2\times2$ factorial experiment (state shifted or not, crossed with chain-end loss weight) shows: shifting the state raises the pass rate from 117/218 to 191/218; up-weighting the loss lowers it (117 to 110; with the state shifted, 191 to 172). In resource terms, this architecture emits 6.12 decision actions per problem on average, takes 0.53 s per problem, and peaks at 833 MiB of inference memory; the comparison model emits 168.0 generated tokens, takes 19.88 s, and peaks at 8,567 MiB. The two systems differ in interface conditions: the comparison model has no execution engine or compiler and must generate the complete proof as free text. We use this comparison as evidence on the structure of the output channel.
The contributions of this paper are as follows. We give, in a controllable domain, the error-product structure of direct long-proof generation and verify it quantitatively. We propose and implement SCOPE, an architecture in which the language model serves as the policy planner, a symbolic execution engine serves as the computing component, and a compiler serves as the text-generation component, reaching an 87.6\% certified pass rate on the test suite. The model's multi-step thinking is endogenous to the architecture and enabled by default: it happens in the discrete decision space over operator chains, produces no natural-language thinking text, and costs 6.12 decision actions per problem on average. Thinking never passes through the output channel, yet the system is not a black box: per-step decision distributions and engine states can be read out directly. A factorial experiment attributes the main information bottleneck of chain decisions to state visibility at decision time rather than to loss allocation. For all certification readings we establish a validity pipeline of version-frozen review plus independent recheck plus reverse verification. We complete a domain-extension validation on a public problem library: the admissible domain is determined by mechanical checks and disclosed layer by layer, model-only readings are reported separately from reference-chain readings, and an independent final audit verifies the numbers. Code, corpus, raw experiment outputs, and the figure-generation pipeline are publicly available at https://github.com/galixiaomaohan/scope-code-data.
\section{Related Work}
The use of language models for theorem proving has evolved from tactic-level prediction to complete proof generation. GPT-f \cite{gptf} and follow-up work train tactic-prediction models in the Lean and Metamath environments; ReProver \cite{reprover} and related retrieval-augmented methods improve proof search with premise selection; the DeepSeek-Prover line \cite{dsp1}, \cite{dsp15} trains proof-specialized models on Lean 4 from large-scale synthetic data, with proof-assistant feedback reinforcement learning entering at V1.5 and evaluation centered on benchmarks such as miniF2F \cite{minif2f} and Lean-Workbook \cite{leanworkbook}; its V1.5 release is the comparison model of this paper. The common form of these methods is what we call the direct-generation paradigm: the model produces proof text or tactic sequences itself. Its failure mode on multi-step numeric propositions is the subject of our first empirical law. Open environments such as LeanDojo \cite{reprover} provide the evaluation infrastructure for this line of work. Another line couples learning with symbolic search: HyperTree Proof Search \cite{htps} explores proof trees under a transformer guide, Magnushammer \cite{magnushammer} replaces hand-tuned premise selection with a learned retriever, Draft-Sketch-Prove \cite{dsp3stage} turns an informal draft into a formal sketch that smaller provers complete, AlphaGeometry \cite{alphageo} pairs a language model with a symbolic deduction engine, and AlphaProof \cite{alphaproof} carries the recipe to competition-level problems. Assistant-style tools bring models inside proof assistants: Baldur \cite{baldur} generates and repairs whole Isabelle proofs from failure feedback, and Lean Copilot \cite{leancopilot} serves tactic suggestions natively in Lean 4; in both, the model remains in the role of producing proof text. Learned neuro-symbolic models of this broad family are an established design space in this journal \cite{tnnlssu}. Relative to these systems, SCOPE's interface differs in three respects: the model never writes proof text or tactics and commits only to operator decisions inside a restricted vocabulary; a template compiler, rather than the model, renders the Lean text; and the symbolic execution engine executes exact integer and rational arithmetic with a checkable certificate at every step, returning the state that each next decision conditions on. What the proof-producing systems above do not provide is a controlled analysis of the failure mechanism that concerns us here, the conjunction structure of long numeric derivations.
Research on tool-augmented language models shows that delegating numeric computation or code execution to external tools improves reliability on the corresponding tasks; representative work includes program-aided language modeling \cite{pal} and self-taught tool invocation \cite{toolformer}, and documented failures of language models on fine-grained character- and symbol-level manipulation \cite{tnnlszhang} sharpen the case for delegation. The difference here is that tool calls are not wholesale delegation (generate a program once, then run it) but are embedded in the step-by-step decision process, with every decision conditioned on the state the tool returns; the tool state itself enters training and inference as model input, and we examine the link between state visibility and decision correctness on that basis. Curriculum learning \cite{curriculum} and constructive data synthesis are widely used in mathematical-reasoning training; the distinctive property of our corpus engineering is auditability: every training row carries an executable reference chain and has passed a mechanically executed leakage check. Work related to our error analysis includes exposure bias and its remedies in sequence generation \cite{schedsampling}, \cite{mixer}, error compounding that deepens with derivation length in compositional tasks \cite{faithfate}, and step-level verification of reasoning chains \cite{letsverify}, with chain-of-thought prompting \cite{cot} as the common substrate. Most of these works observe the phenomenon indirectly, through end-to-end metrics on natural-language reasoning tasks; we reduce it, in a certified setting, to a computable structure, namely a Boolean conjunction of proof validity over content elements, which yields a definite power-law relation between pass rate and per-integer accuracy and admits direct examination under controlled conditions.
\section{Method}
This section describes the components of SCOPE (Fig. 1). The system takes a mathematical proposition written in a domain-specific language as input and outputs a proof certified by the Lean compiler. Processing has four parts: the discrete representation of propositions and the operator vocabulary (Section III-A), the symbolic execution engine (Section III-B), the chain-planning model (Section III-C), and the proof compiler (Section III-D); the review protocol is given in Section III-E.
\begin{figure*}[!t]
\centering
\includegraphics[width=\textwidth]{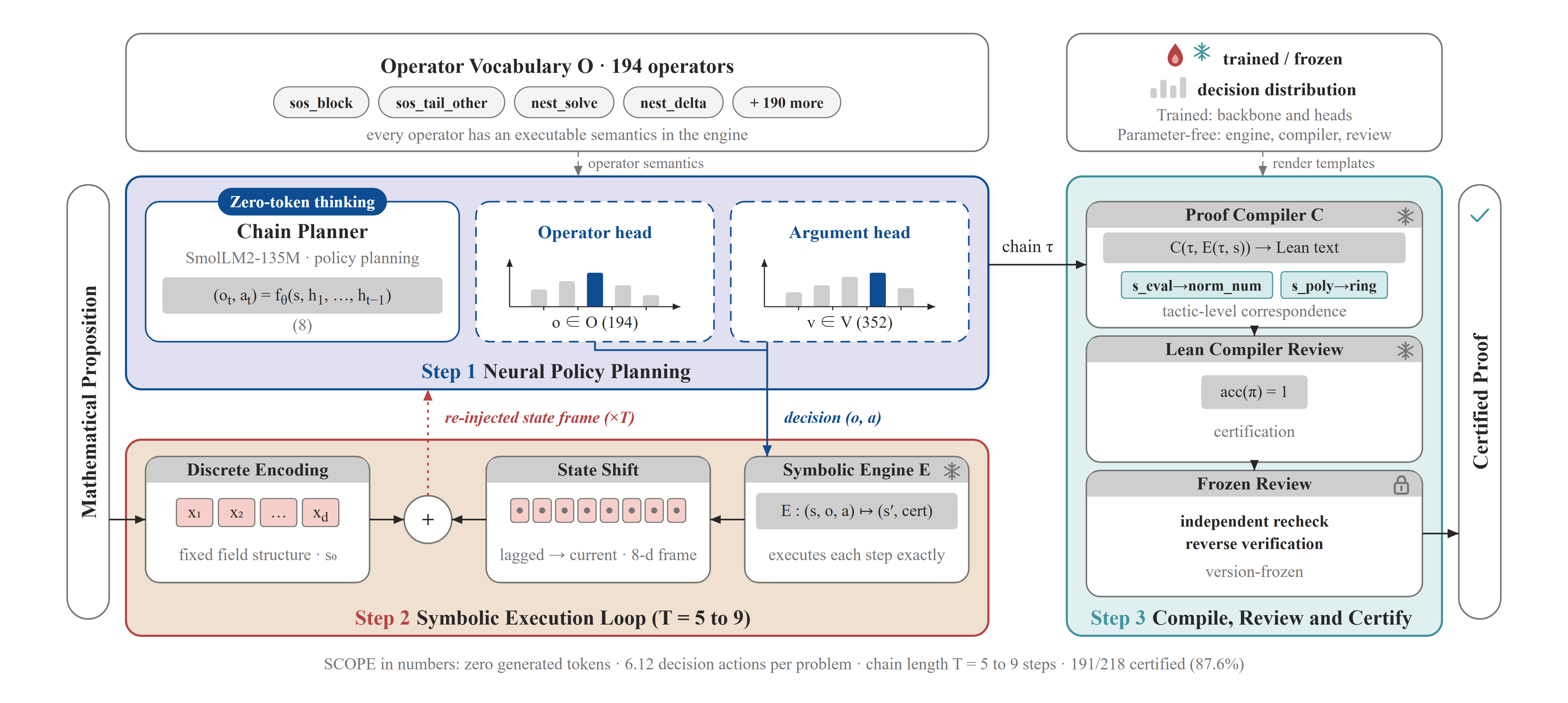}
\caption{The SCOPE architecture and data flow. A mathematical proposition enters a latent decision loop through discrete encoding: the chain planner, with its operator and argument heads, handles perception and routing and commits to one decision per step inside the operator vocabulary, which supplies both operator semantics and rendering templates; the execution engine executes each step exactly, and the state frame, after the state shift, feeds back into the decision loop, so multi-step thinking is carried endogenously by the loop with zero generated tokens (labeled Zero-token thinking in the figure). The chain and its arguments go to the proof compiler, which renders Lean text from the templates; the Lean compiler checks it, a frozen review program performs an independent recheck and reverse verification, and a certified proof is output. The trained components are the backbone and the two heads (flame); the engine, the compiler, and the review program are parameter-free (snowflake), and the review program is version-frozen (lock). The bottom line of the figure gives the operating profile: zero generated tokens, 6.12 decision actions per problem, chain length T = 5 to 9 steps, and 191/218 certified (87.6\%).}
\end{figure*}
\subsection{Task Definition and Operator Vocabulary}
The problem domain covers three families of elementary discrete mathematics: induction identities, polynomial inequalities, and elementary number-theoretic divisibility. Each proposition is produced constructively by a parameterized generator together with a reference solution chain and a reference proof; generation guarantees solvability, and the reference solution is verifiable by mechanical execution. A proposition is given as a state: the objects it mentions (variables, polynomial coefficients, moduli) appear as a fixed field structure rather than as natural language. Solving is modeled as an operator chain, where each step consists of an operator identifier and a list of concrete arguments, and the chain ends with an explicit terminator. All legal operators form a finite vocabulary; the initial release contains 194 operators, extended to 220 for the public library (Section X). The vocabulary grows in successive batches: a new batch of operator rows enters incremental training only if it passes a zero-regression gate on existing capability. Every operator has an executable semantics in the engine. The vocabulary is organized by problem family, for example the nested-induction family, the lemma-tower family, and the sum-of-squares decomposition family; operators of one family share argument structure and state-update form. The model never touches Lean tactic names; the correspondence at tactic level is handled by the compiler.
For later reference, we fix the formal notation as follows.
\emph{Definition 1 (problem state):} a problem state is an integer vector with a fixed field structure,
\begin{equation}s = (x_1,\ldots,x_d)\in\mathbb{Z}^{d},\end{equation}
where the fields correspond to the objects of the proposition (variables, polynomial coefficients, moduli), and one problem-type family shares one field structure.
\emph{Definition 2 (operators and solution chains):} the legal operators form a finite vocabulary O (194 initially, 220 after the public-library extension); each operator $o\in O$ carries an argument vector a of operator-specific shape; one decision is the pair (o, a). A solution chain is a finite sequence
\begin{equation}\tau = \bigl((o_1,a_1),\ldots,(o_T,a_T)\bigr),\end{equation}
where T is the chain length, and the chain terminates with an end symbol.
\emph{Definition 3 (execution engine and certification):} the execution engine is a deterministic map $E:(s,o,a)\mapsto(s',\mathrm{cert})$ that returns a unique state $s'$ and an execution certificate cert for identical inputs. Write $\mathrm{acc}(\pi)=1$ if proof text $\pi$ passes certification, that is, it compiles in Lean and the compiled result depends on no unproven assumption; every certified pass rate reported in this paper is the mean of acc over the corresponding problem set.
Definitions 1 to 3 fix the formal objects of the system, and the solving goal can now be stated: given a problem state s, find a solution chain whose compiled proof passes certification, that is,
\begin{equation}\text{find } \tau \text{ such that } \mathrm{acc}\bigl(C(\tau,s)\bigr) = 1,\end{equation}
where C is the proof compiler (Section III-D, formalized in (12); here C(${\tau}$, s) abbreviates C(${\tau}$, E(${\tau}$, s))) and acc is the certification indicator of Definition 3. This search cannot be carried out by enumerating chains: the number of candidate combinations per step is the product of the vocabulary size and the value-space size (220 and 352 in the public-library configuration; the initial release has 194 operators), chains run 5 to 9 steps, and the chain space exceeds the fifth power of that product. The role of the chain-planning model (Section III-C) is to locate the correct chain in this space, step by step, under a learned conditional distribution.
Table I maps the problem-domain layers onto the composition of the vocabulary. The vocabulary grows in batches, each covering a problem type or one domain extension; every batch has executable semantics in the engine (Section III-B), and the vocabulary carries a per-operator admissibility matrix over multiplier rows that restricts the legal argument set of each decision.
\begin{table}[H]
\caption{Problem-domain layers and the composition of the operator vocabulary. The initial release has 194 rows (first four segments) and the public-library extension raises the count to 220 (Section X); per-segment counts are measured from the system's vocabulary registry.}
\centering
\footnotesize
\setlength{\tabcolsep}{3pt}
\begin{tabular}{>{\centering\arraybackslash}p{0.23\columnwidth}c>{\centering\arraybackslash}p{0.54\columnwidth}}
\toprule
\textit{Vocabulary segment} & \textit{Rows} & \textit{Covered problem types and introduction batch} \\
\midrule
Base segment & 161 & Core operators and skeleton rows of the initial three domains (induction identities, polynomial inequalities, number-theoretic divisibility) \\
Deep-chain extension segment & 17 & Closed-form operators of the deep-chain corpus: sum-of-squares closed forms, geometric and summation closed forms, block closing, partial fractions, and high-tower combinations \\
Construction-curriculum segment & 7 & Construction curricula: induction recursion, ring identities, sum-of-squares decomposition, and divisibility witnesses \\
Deep-chain family segment & 9 & Nested-induction chain construction (difference-equation solving, folding, telescoping check, closed-form closing) and sum-of-squares chain construction (single-variable blocks, cross blocks, three closing types) \\
Substitution and evaluation segment & 4 & Algebraic evaluation, modular evaluation, and variable substitution \\
Routing segment & 9 & Route operators from public-library problem statements to existing families \\
Extension segment & 13 & Nine extension families: goal-shape splitting, quantifier introduction, existence witnesses, predicate unfolding, radical reduction, symbolic exponentiation, factorials, library-name entry, and chained relations \\
Total & 220 & N/A \\
\bottomrule
\end{tabular}
\end{table}
\subsection{Symbolic Execution Engine}
The execution engine is a set of deterministic symbolic-computation programs, one execution function per operator. Given the current state and the arguments, the function performs the computation and returns two things: the numeric intermediate result, and a mechanically checkable execution certificate (for example, an algebraic identity check that passed). Taking the sum-of-squares decomposition family as an example, each step removes one square component from the goal expression and returns what remains, which we call the residual; its update rule is
\begin{equation}R_{\mathrm{out}} = R_{\mathrm{in}} - m\,\ell^{2},\end{equation}
where $R_{\mathrm{in}}$ and $R_{\mathrm{out}}$ are the residuals before and after the step, m is the coefficient of the removed square term, and ${\ell}$ is the corresponding linear form. The residual enters the model's state frame as an eight-dimensional numeric vector, re-injected from the engine at every step (Sections III-C and VIII). All engine computation uses exact integer and rational arithmetic and contains no floating-point error; the engine is implemented in C++/CUDA, is bit-aligned with the training-side reference implementation, and runs more than five times faster than it. The final step of a chain must classify the final residual ${\rho :}$
\begin{equation}\begin{aligned}c(\rho) &= 0\ (\rho = 0);\quad c(\rho) = \mathrm{square}\ (\rho \in \{4,9,16,25,36,49\});\\ c(\rho) &= \mathrm{other}\ (\text{otherwise})\end{aligned}.\end{equation}
This correspondence holds for every sum-of-squares row of the corpus and every test-suite problem of the family (799/799 and 181/181) and underlies the factorial analysis of Section VIII. Two design constraints govern the engine-model interface. First, the arguments fed to the engine take reference values during training and model predictions at inference; if a predicted argument is illegal (for example, outside its domain), the engine rejects execution and the step is scored as an error. Second, the engine returns a fixed-dimensional state for every operator, so the states of different operators can be concatenated within one frame structure.
For the mainstay operator of the sum-of-squares family (single-variable block absorption), executing (4) inside the engine must satisfy an exact absorption criterion: one square component is removed if and only if the residual decomposes into that square term and a remainder that contains no nonzero term in any variable supporting the removed linear form, that is,
\begin{equation}\begin{aligned}R_{\mathrm{in}} &= m\,\ell^{2} + R_{\mathrm{out}},\\ &\text{and the coefficients of the variables}\\\ &\quad\text{of }\ell\text{ in }R_{\mathrm{out}}\text{ are all zero}\end{aligned}.\end{equation}
The two conditions together make the removal exact; arguments that fail the criterion (for instance m not a positive integer, or a remainder that still contains terms in variables supporting ${\ell )}$ are rejected by the engine. The criterion is verified mechanically on a finite test grid:
\begin{equation}\begin{aligned}R_{\mathrm{in}}(k) &= m\,\ell(k)^{2} + R_{\mathrm{out}}(k),\\ &\text{for all } k \text{ on the verification grid}\end{aligned}.\end{equation}
The grid is determined by the variable domains and a degree bound; passing the verification issues the execution certificate of the step, which is written to disk together with the chain (Section III-G).
Table II lists representative operators of each problem family and their execution logic. Operator names are the identifiers of the implementation (family-prefixed in the released code); for each family we show one or two operators that carry its main computation; full specifications of all operators ship with the code.
\subsection{Chain-Planning Model}
The chain-planning model uses SmolLM2-135M \cite{smollm2}, a transformer decoder \cite{transformer}, as the backbone and adds two output heads: the operator-selection head gives, over the vocabulary, the distribution of the next operator, and the argument-prediction head gives that operator's arguments, predicted in a discretized value space (352 candidates). Writing the problem state as s and the representation of the executed steps as $h_1,\ldots,h_{t-1}$ (each a 1,152-dimensional last-layer state of the backbone), the decision at step t is
\begin{equation}(o_t,a_t) = f_{\theta}(s, h_1, \ldots, h_{t-1}),\end{equation}
where o${_{t}}$ is the operator choice, a${_{t}}$ the arguments, and ${\theta}$ the model parameters. Decisions proceed autoregressively until the model emits the terminator or the step budget is reached.
The two output heads expand $f_{\theta}$ into discrete distributions. Write the decision frame as $F_t=(s,h_1,\ldots,h_{t-1})$; the final configuration extends this frame with the engine state (Section VIII, (19), and Algorithm 2); the operator head gives a distribution over the vocabulary O, and the argument head a distribution over the value space $V$ with $|V| = 352$:
\begin{equation}\begin{aligned}P(o_t = o \mid F_t) &\propto \exp\bigl(z_o(o)\bigr),\\ P(a_t = v \mid o_t, F_t) &\propto \exp\bigl(z_a(v)\bigr)\end{aligned},\end{equation}
where $z_o$ and $z_a$ are the logits of the two heads on the decision frame, and the legal value set of each operator is restricted by the admissibility matrix carried with the vocabulary (Table I). At inference, the highest-probability item is taken at every step.
The training objective (Algorithm 1) is the cross-entropy of operator selection and argument prediction, supervised by the reference chain $(o_t^{*},a_t^{*})$:
\begin{equation}\mathcal{L} = \mathcal{L}_{\mathrm{op}}(o_t, o_t^{*}) + \tfrac{1}{24}\,\mathcal{L}_{\mathrm{val}}(a_t, a_t^{*}),\end{equation}
where the argument term is scaled by the number of value slots per argument, namely 24. The chain-end weighting configuration gives the operator cross-entropy at the terminating step ($t=T$) an extra weight, bringing its total weight to ${\lambda  (}$Experiment 4 uses $\lambda=5$):
\begin{equation}\mathcal{L}_{W} = \mathcal{L} + (\lambda - 1)\,\mathbf{1}[t = T]\,\mathcal{L}_{\mathrm{op}}.\end{equation}
All main configurations (Experiments 1 to 4) share one training configuration: AdamW optimization, 6,000 steps, batch size 6, learning rate 4${\times 10^{-5}}$, random seed 20260923, and the same batch-order file (whose checksum is identical across configurations), so that any difference between configurations is attributable to a single variable. The training dynamics of the final configuration (loss components and gradient norms over the 6,000 steps) appear in Fig. A1. An earlier version of the model also carried a proof-text generation head (used in Experiments 1 and 2 of Section VI); it was removed following the findings of Section VII.
\begin{table*}[!t]
\caption{Execution logic of representative operators. The execution semantics and certificates are the actual checking procedures inside the engine; the rendered tactics are the Lean tactics the compiler expands for that step.}
\centering
\scriptsize
\setlength{\tabcolsep}{3.5pt}
\begin{tabular}{>{\centering\arraybackslash}p{0.15\textwidth}>{\centering\arraybackslash}p{0.34\textwidth}>{\centering\arraybackslash}p{0.30\textwidth}>{\centering\arraybackslash}p{0.13\textwidth}}
\toprule
\textit{Operator (family)} & \textit{Execution semantics} & \textit{Execution certificate} & \textit{Rendered tactics} \\
\midrule
sos\_block (sum-of-squares) & Deducts one single-variable square component from the current residual per (4) & Grid identity (7) plus absorption criterion (6); m a positive integer & positivity \\
sos\_tail\_other (sum-of-squares) & Classifies the final residual per (5) and closes the chain & Closed-chain identity: the goal equals the sum of deducted squares plus the final residual & norm\_num, nlinarith only \\
nest\_solve (nested induction) & Solves one difference equation from increment and initial value, yielding a polynomial closed form & Quadruple check: exact degree, initial value, telescoping grid, degree ladder & rw and ring inside induction \\
nest\_delta (nested induction) & Verifies that the closed form satisfies $F(k+1)-F(k)=S(k)$ & Pointwise identity of the forward-difference grid & Certificate-level check; emits no tactic lines \\
s\_poly (routing) & Recognizes the statement as a polynomial identity and invokes ring-law verification & Ring-law recomputation & ring \\
s\_eval (routing) & Recognizes the statement as numeric evaluation and computes step by step & Exact arithmetic recomputation & norm\_num \\
sqrt\_iso (extension families) & Gives exact values for radicals of perfect-square closed forms; squares one-sided radical inequalities under side conditions & Grid check of side conditions & rw (radical equivalence theorems) \\
intro\_binder (extension families) & Strips the leading universally quantified binder and passes the body to existing routes & Structural check & intro \\
\bottomrule
\end{tabular}
\end{table*}
\begin{algorithm}[H]
\caption{Training the chain-planning model}
\label{alg:training}
\begin{algorithmic}[1]
\REQUIRE training corpus $D$ (the two layers of Section~IV-B, with reference chains $(o^{*},a^{*})$ and per-step ground-truth engine states); batch-order file; 6,000 training steps; batch size 6; learning rate $4\times10^{-5}$; decision-frame state timing (original or shifted, per the configuration). Convention: during training, the state and arguments in the decision frame take reference values (Section~III-B).
\STATE initialize parameters $\theta$ (SmolLM2-135M backbone plus operator-selection and argument-prediction heads)
\FOR{step $= 1$ \TO $6{,}000$}
  \STATE draw a mini-batch $\{(s,o^{*},a^{*})\}\subset D$ following the batch-order file
  \STATE for each sample, read the decision frame step by step per the configuration's state timing; compute $(o_t,a_t)\gets f_{\theta}(\cdot)$
  \STATE $L \gets L_{\mathrm{op}}(o_t,o_t^{*}) + (1/24)\,L_{\mathrm{val}}(a_t,a_t^{*})$ \COMMENT{(10)}
  \STATE if chain-end weighting enabled (Experiments~4B/C, $\lambda=5$): $L \gets L + (\lambda-1)\,\mathbf{1}[t=T]\,L_{\mathrm{op}}$ \COMMENT{(11)}
  \STATE $\theta \gets \mathrm{update}(\theta,\nabla L)$
\ENDFOR
\ENSURE trained parameters $\theta$ and the training log (steps, losses, and checksums).
\end{algorithmic}
\end{algorithm}
\subsection{Proof Compiler}
Proof text is produced by a compiler: given an operator chain, the compiler expands each step into Lean tactic sequences under the operator's template rules, and every number in the output comes from the engine's actual computation. The compiler calls no language model and reads no reference proof text. Its correctness rests on two verifications: on all 2,399 corpus rows, the compiled output matches the reference proof byte for byte (2,399/2,399); on the test suite, feeding the reference chains yields proofs that all certify (218/218). Consequently, whenever the model's chain is correct, the compiled proof certifies, and the certified pass rate of the system is determined entirely by the quality of chain decisions.
Compilation is a deterministic map. Writing E(${\tau}$, s) for the step-by-step execution results (the state and certificate sequence) of chain ${\tau}$ started from state s, the proof text is
\begin{equation}\pi = C\bigl(\tau, E(\tau, s)\bigr).\end{equation}
The map C has no free parameters and calls no language model: template rules fix the tactic skeleton, and all numbers come from the actual computation in E(${\tau}$, s). Since C reproduces the reference proofs byte for byte on the corpus (2,399/2,399), the search problem of (3) reduces entirely to the chain-decision problem.
\begin{algorithm}[H]
\caption{SCOPE solving one problem}
\label{alg:solving}
\begin{algorithmic}[1]
\REQUIRE problem state $s$; operator vocabulary $O$; execution engine $E$; proof compiler $C$; step budget $T_{\max}$ (set beyond the corpus chain lengths of 5 to 9 steps); decision-frame state timing (original or shifted).
\STATE $h\gets\emptyset$; $t\gets 1$
\WHILE{$t \leq T_{\max}$}
  \STATE read decision frame $F_t=(s,h_1,\ldots,h_{t-1})$, plus $R_{\mathrm{out}}^{(t-1)}$ in the shifted timing \COMMENT{(19)}
  \STATE $(o_t,a_t)\gets f_{\theta}(F_t)$ \COMMENT{(8)}
  \STATE if $o_t$ is the terminator then break
  \STATE $(s,\mathrm{cert})\gets E(s,o_t,a_t)$; $h_t\gets(o_t,a_t)$ \COMMENT{(4)}
  \STATE $t\gets t+1$
\ENDWHILE
\STATE $\tau\gets(h_1,\ldots,h_{t-1})$; $\pi\gets C(\tau, E(\tau, s))$
\ENSURE the operator chain $\tau$, the proof text $\pi$, and per-step records (decision distributions, arguments, engine states, and certificates).
\end{algorithmic}
\end{algorithm}
\subsection{Review Protocol}
Review is performed by an independent automatic program: a candidate proof is compiled in the Lean environment; the program requires compilation to succeed, prints the proof's axiom dependencies, and rejects any proof whose dependencies include sorryAx (Lean's axiom for unproven goals); a second pass recompiles the proof with autoImplicit and relaxedAutoImplicit disabled and treats any failure as rejection. The program stayed version-frozen for the entire experimental period (anchored by an MD5 checksum, given in Appendix B), and every experiment was checksummed once before and once after review. To rule out bias introduced by review itself, a dual-protocol recheck is executed: at least 60 accepted proofs, drawn in strata, are re-reviewed one by one through an independent path; in the reverse direction, 12 corrupted proofs of each of two kinds (final segments truncated; proofs of other problems substituted) are submitted, and the program is required to reject them. Across all experiments the two protocols show zero discrepancies, and reverse verification rejects all cases.
\subsection{Auditability}
The design makes the entire solving process inspectable step by step. The model's output is an operator chain inside a restricted vocabulary: each step consists of an operator identifier and concrete arguments, whose legality can be checked one item at a time against the engine's semantics. This property contrasts with the direct-generation paradigm: the intermediate steps of a free-text proof admit no mechanical inspection, and evaluation must rely on the final compilation result. In this architecture, any output can be replayed as a complete decision sequence, and an error can be pinned to a specific step. Auditability at the computing layer comes from the determinism of the engine. The engine executes every step in exact integer and rational arithmetic and returns the numeric state plus a mechanically checkable execution certificate; for example, the correspondence between the final class and the residual value holds row by row on the sum-of-squares rows of the corpus and the test suite (799/799 and 181/181), so the correctness of this stage can be verified independently of the model. Proof text is generated by the compiler from templates; it matches the reference proofs byte for byte on all 2,399 corpus rows and certifies on all 218 test problems given the reference chains, showing that the text stage contains no free parameters. Auditability at the review layer rests on three mechanisms: the review program is version-frozen (MD5-anchored, checked before and after grading in every experiment); accepted samples are re-reviewed through an independent path in strata (66/66 zero discrepancies, replicated across two configurations); reverse verification rejects all corrupted proofs (12/12 for both truncation and substitution). Two zero-capability controls (the untuned backbone and a rule-based solver, 1,080 applicable problems each with zero passes) further rule out earning certified scores by guessing or pattern matching. Auditability at the experiment layer takes the form of configuration control and record completeness: all configurations share the batch-order file (identical checksums) and random seeds; the differing variable of each configuration is registered in the design file in advance; every results file carries a checksum ledger. A direct payoff of auditability is error analysis at decision-step granularity: the 101 first errors of Section VIII all localize to the final step of the chain, which presupposes the symbolic decision sequence; attribution at this granularity cannot be obtained directly from free-text output.
\subsection{Process Auditability and Observability of Latent Thinking}
The intermediate process consists of explicit symbolic structures and can be inspected item by item. For every problem, the system writes to disk: the model's complete operator chain, the arguments of every step, the residual and execution certificate returned by the engine, the proof text produced by the compiler, and the review program's verdict with its reason (the compilation result and the axiom dependencies). Errors can therefore be pinned to specific steps: Section VIII localizes all 101 errors to the final step of the chain and reports the predicted probability at each erroneous step (55 of them at 0.9 or above). Step-level analysis of this kind lacks an operational counterpart in the free-text generation paradigm: text output offers only an aggregate right-or-wrong signal and surface statistics.
In the final configuration of this paper, multi-step thinking is the default behavior granted directly by the training objective; no prompt or mode switch triggers it. The entire solving process consists of step-by-step decisions over the operator chain, and the model produces no natural-language output: on average 6.12 discrete decision actions per problem (Section IX), with the proof text compiled separately. Because the carrier of decision is a discrete symbolic structure, the thinking process is observable from outside, and the observation does not rely on the model's self-report: the operator-selection distribution, the predicted arguments, and the engine-returned state of every step can be read and tallied directly. The hidden state of the decision frame is equally readable: in low-dimensional projection it organizes by problem family (nearest-neighbor purity 0.999), the frames of one problem advance continuously along the chain (Fig. 2), and the vectors that enter the projection are frame-for-frame identical to the hidden states that the output heads actually consume. By comparison, the observability of chain-of-thought comes at the price of generated tokens, and natural language produced as an explanation can disagree with the model's actual reasons for deciding; what we observe here is the decision itself.
Every stage of the system can be re-verified by a third party. The review program is version-frozen and carries a checksum; accepted samples are re-reviewed through an independent path (stratified sampling, zero discrepancies); reverse-constructed corrupted proofs are all rejected (Section III-E); training configurations, batch order, and random seeds are registered together with the checksums of all reading files, so any reading can be recomputed from the released material; proof compilation is deterministic and fully replayable from the operator chain (2,399/2,399 byte-identical on the corpus, Section III-D); the isolation check between corpus and test suite is a two-level mechanical scan whose program ships with the code. Auditability also makes controlled intervention possible. The factorial experiment of Section VIII directly edits the state quantity inside the decision frame and retrains comparison configurations, thereby attributing a 74-problem change in pass rate to a single variable; frame-level intervention of this kind presupposes a structured process. We will release the code, corpus, review programs, and per-problem records (including per-step decision distributions), so that all experimental readings can be re-checked outside the system.
\subsection{Worked Examples: From Problem State to Certified Proof}
This section presents the input, the decision chain, and the output of the system on four real problems. Two come from the main suite (the configuration of Section VIII) and two from the public-library final exam (Section X); the chains and proof texts are the system's actual records on disk, unedited.
\begin{figure*}[!t]
\centering
\includegraphics[width=\textwidth]{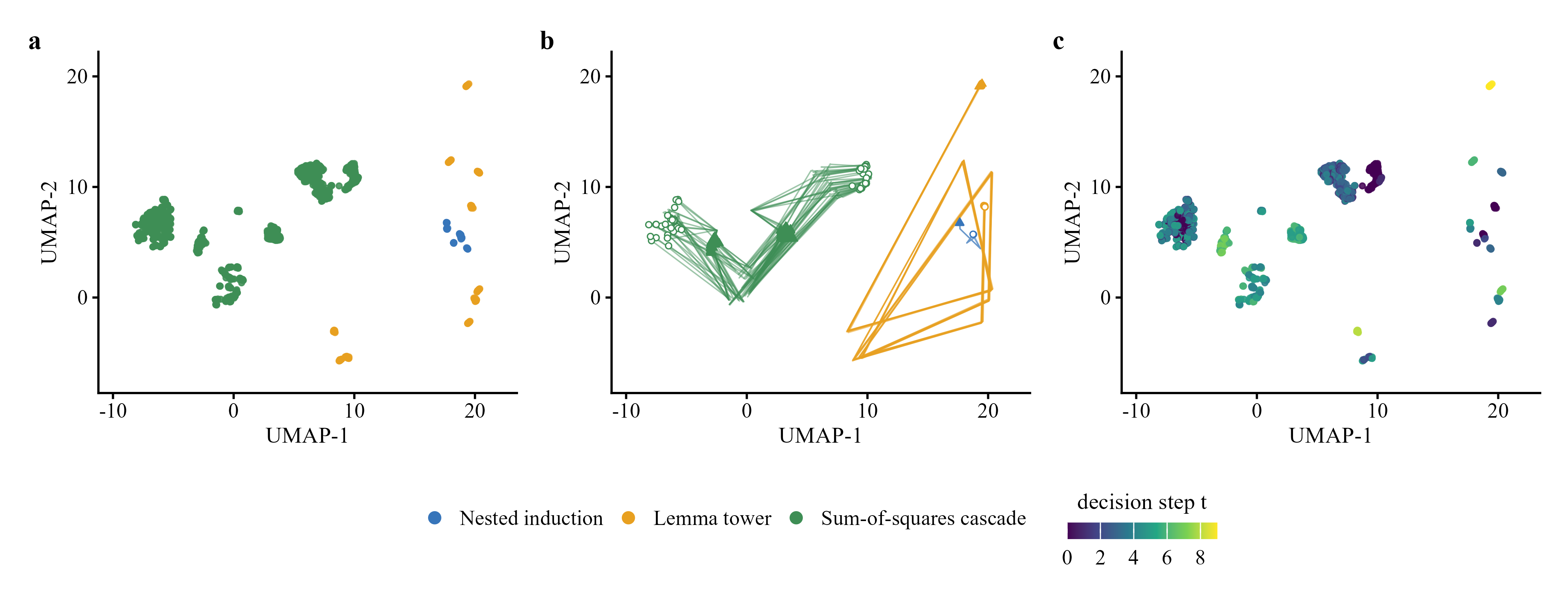}
\caption{Low-dimensional projection of decision-frame hidden states (218 problems, 1,552 frames including the terminator frame of each chain, uniform manifold approximation and projection (UMAP) \cite{umap}). (a) All 1,552 frames, colored by problem family; the three families separate almost perfectly in the embedding (nearest-neighbor purity 0.999 against a random baseline of 0.659). (b) The 80 sampled chains as step-ordered trajectories; open markers mark each chain's first frame and solid markers its last. (c) All frames, colored by decision step. Frames of the same problem advance along the chain: adjacent-frame cosine similarity is 0.607 on average (5 to 95\%: 0.431 to 0.790) against 0.154 for random frame pairs, all measured in the full 1,152-dimensional hidden space, and the vectors that enter the projection are frame-for-frame identical to the hidden states consumed by the output heads (maximum deviation 0). Thinking rides on a discrete decision sequence whose internal state can be read out directly, without relying on the model's self-report.}
\end{figure*}
The sum-of-squares example shows the residual mechanism in full. The problem is the multivariate polynomial inequality $v^{2}+2u^{2}+2w^{2}+z^{2}-2yv+y^{2}+2x^{2}+4v-12u-4w-4z-4y-8x+57\geq-7$. The model's chain has six steps: cross-block absorption handles the square containing the cross term $-2yv$; four single-variable block absorptions remove four square terms in succession; and the closing step classifies the final residual. Table III lists, step by step, the square component removed at each step and the proof line rendered from it. The final residual is 28, of the other class per (5); the closing accordingly emits a constant-nonnegativity proof line and then closes all goals with a nonlinear-arithmetic tactic. The proof was certified by the review program.
The nested-induction example is a closed-form problem: the proposition is $\forall n:\mathbb{N},\ f n = 4-12n+n^{2}+12n^{3}+4n^{4}$, with f defined by a difference recurrence. The model's five-step chain solves the inner difference equation for the inner closed form, folds it into the outer increment, solves the outer difference equation, runs a telescoping check confirming that the closed form matches the increment pointwise (Table II), and closes with the closed form together with its class. The rendered proof is a two-layer induction structure: an induction first establishes the closed form of the inner recurrence, then an induction on the outer goal uses the previous step's result, and ring normalization finishes the proof. The proof was certified by the review program.
The two public-library examples come from the Lean-Workbook final exam. For the algebraic identity $x^{5}-y^{5}=(x-y)(x^{4}+x^{3}y+x^{2}y^{2}+xy^{3}+y^{4})$, the chain has two steps: a routing reduction into the polynomial route, rendered as the two tactics norm\_num and ring. For the number-theory evaluation $(29\times31+37-41)\bmod 4=3$, the chain has the same two-step shape into the evaluation route, rendered as a single line of norm\_num. Both problems certify. Full records of all four examples (per-step decision distributions, engine states, and certificates) ship with the code.
\begin{table*}[!t]
\caption{Step-by-step computation of a sum-of-squares problem (actual record on disk). Each step removes one square component per (4); every number in the rendered line comes from engine computation; the closing row gives the final residual 28 (other class, per (5)).}
\centering
\scriptsize
\setlength{\tabcolsep}{3.5pt}
\begin{tabular}{c>{\centering\arraybackslash}p{0.13\textwidth}>{\centering\arraybackslash}p{0.23\textwidth}>{\centering\arraybackslash}p{0.41\textwidth}}
\toprule
\textit{Step} & \textit{Operator} & \textit{Component removed, $m\,\ell^{2}$ per (4)} & \textit{Rendered proof line (engine-computed values)} \\
\midrule
1 & sos\_cross (cross block) & $1\,(-v+y-2)^{2}$ & {\ttfamily have h1 : (0:$\mathbb{Q}$) $\leq$ 1 * (- v + y - 2) \textasciicircum{} 2 := by positivity} \\
2 & sos\_block & $2\,(w-1)^{2}$ & {\ttfamily have h2 : (0:$\mathbb{Q}$) $\leq$ 2 * (w - 1) \textasciicircum{} 2 := by positivity} \\
3 & sos\_block & $2\,(u-3)^{2}$ & {\ttfamily have h3 : (0:$\mathbb{Q}$) $\leq$ 2 * (u - 3) \textasciicircum{} 2 := by positivity} \\
4 & sos\_block & $1\,(z-2)^{2}$ & {\ttfamily have h4 : (0:$\mathbb{Q}$) $\leq$ 1 * (z - 2) \textasciicircum{} 2 := by positivity} \\
5 & sos\_block & $2\,(x-2)^{2}$ & {\ttfamily have h5 : (0:$\mathbb{Q}$) $\leq$ 2 * (x - 2) \textasciicircum{} 2 := by positivity} \\
6 & sos\_tail\_other (closing) & $\rho=28$ & {\ttfamily have hrho : (0:$\mathbb{Q}$) $\leq$ (28:$\mathbb{Q}$) := by norm\_num; nlinarith only [h1, h2, h3, h4, h5, hrho]} \\
\bottomrule
\end{tabular}
\end{table*}
\section{Experimental Setup}
\subsection{Pre-Control on the Carrier of Latent Thinking}
A precondition of this architecture concerns the form the carrier of thinking takes inside the latent space: if the intermediate states of multi-step reasoning are passed between the model's hidden states as continuous vectors, can later steps read them out reliably? Latent-carrier designs of this kind have recently been explored as alternatives to token-level reasoning chains \cite{coconut}, \cite{recurrentdepth}; whether the carrier survives across steps is a property to be tested rather than assumed. A controlled experiment under a unified protocol (same backbone, same number of training steps, two random seeds) gives a negative answer for continuous feedback. As Table IV shows, pure hidden-state feedback reaches a readout accuracy of 0.01 to 0.06 beyond two hops; the continuous-register scheme averages 0.0246 over all depths; value-format discretization reaches 10\% to 15\%. The only configuration that reads out fully at a depth of 16 hops is key-token discretization combined with curriculum training, which scores 1.0000 at every depth $k=1$ to $16$ on formal evaluation. Neither ingredient suffices alone: curriculum by itself reads out at roughly zero, and key format alone needs roughly three to four times the training volume to bootstrap. The external-judge reference group (0.973 to 0.984) shows that the discrimination step itself is feasible; the state is what fails to survive the carrier.
\begin{table}[!t]
\caption{Carrier comparison for multi-hop readout in latent space (unified protocol: same SmolLM2-135M backbone, same number of training steps, two random seeds; readings are readout accuracy; GRU denotes gated recurrent unit).}
\centering
\footnotesize
\setlength{\tabcolsep}{3pt}
\begin{tabular}{>{\centering\arraybackslash}p{0.32\columnwidth}>{\centering\arraybackslash}p{0.34\columnwidth}>{\centering\arraybackslash}p{0.20\columnwidth}}
\toprule
\textit{Carrier configuration} & \textit{Readout accuracy} & \textit{Verdict} \\
\midrule
Pure hidden-state feedback (continuous vectors) & 0.01 to 0.06 beyond two hops (tail mean 0.0223) & Failure \\
Value-format discretization (values only) & 0.10 to 0.15 & Failure \\
Curriculum training (no key format) & About 0 & Failure \\
Key-token discretization (no curriculum) & Slow bootstrap; needs roughly 3 to 4 times the training volume & Partial \\
Continuous registers & Mean 0.0246 over all depths & Failure \\
External judge references (Markov model/GRU) & 0.973 to 0.984 & Judge feasible; no carrier specificity \\
Key-token discretization + curriculum training & 1.0000 at every depth $k=1$ to $16$ & Full readout at 16 hops \\
\bottomrule
\end{tabular}
\end{table}
The comparison readings of Table IV come from the discretization line of latent chain-of-thought, whose main configuration (v15 in our notation) is decomposed in Table V. That line uses a dynamic multi-hop task as its carrier: given one member of an operator family (orthogonal, permutation, or cyclic matrix families), the model applies it to a latent state step by step and reads out the terminal state after T steps. The depth curriculum of v15 advances as a ladder: the depth measured by the gate probe rises monotonically from 1 to 16, with all fifteen openings falling between training steps 11,000 and 17,500 (Fig. A2), and the opening of the depth gate was independently re-verified. Depth sampling of training episodes stayed shallower than the gate depth throughout this window (241 of 377 episodes remained at $T=1$; Fig. A2(c)), so the ladder curve measures the trajectory of the gate, not the training distribution itself. After the ladder was fully traversed from $T=1$ to $16$, the discrete decision carrier of all subsequent experiments in this paper was in place; the comparison readings in the table share the unified protocol of Table IV.
This set of comparisons establishes a necessary condition behind the subsequent design: state that must persist across steps inside the model's own latent activations has to be re-anchored to discrete symbols (keys in token format) to survive. The operator vocabulary of Sections III-A to III-D presupposes this discrete anchor for the model's internal state. The engine-computed state follows a different route: it is re-injected into the decision frame from the engine at every step and never has to survive in latent space (Sections III-B and VIII). The measured depth of 16 hops indicates that the carrier does not bottleneck the task depths of this paper.
\subsection{Training Corpus}
\begin{table}[H]
\caption{Mechanism decomposition of the discretization experiment (v15 configuration). The discretization of keys and values, the feedback channel, and the depth protocol are independent and mutually supporting; the control readings come from arm comparisons under the unified protocol (same source as Table IV).}
\centering
\footnotesize
\setlength{\tabcolsep}{3pt}
\begin{tabular}{>{\centering\arraybackslash}p{0.14\columnwidth}>{\centering\arraybackslash}p{0.48\columnwidth}>{\centering\arraybackslash}p{0.24\columnwidth}}
\toprule
\textit{Mechanism component} & \textit{Treatment in v15} & \textit{Related control readings} \\
\midrule
Task carrier & Dynamic multi-hop: one operator family (orthogonal, permutation, or cyclic matrices) acts on the latent state step by step; the terminal state is read out after T steps & N/A \\
Value discretization & Codebook quantization: the latent state snaps to the nearest codeword; readout and feedback both use the dequantized codeword & Continuous-vector feedback: 0.01 to 0.06 beyond two hops \\
Key discretization & Keys represented in token-vocabulary format & Values discretized without keys: 0.10 to 0.15 \\
Depth protocol & Curriculum ladder: training depth T rises stepwise from 1 to 16, advancing after each level passes the readout gate & Key format without curriculum: roughly 3 to 4 times the training volume \\
Depth outcome & Ladder fully traversed from $T=1$ to $16$; gate openings independently re-verified & Continuous registers: mean 0.0246 \\
\bottomrule
\end{tabular}
\end{table}
The training corpus has two layers. The shallow corpus contains 9,366 proposition-proof pairs with chains of one to three steps, covering the early problem families. The deep corpus contains 2,399 proposition-proof rows built by three deep-chain generator families: the nested-induction family (chain length 5), the lemma-tower family (chain length 9), and the sum-of-squares decomposition family (chain length 5 to 7); the overall chain-length distribution is 1,136 rows at 5 steps, 330 at 6, 133 at 7, and 800 at 9, averaging 6.58 steps. The deep-chain corpus is designed so that multi-step planning genuinely occurs during training: the average chain of the early shallow corpus is only about 1.77 steps, too short to support a study of multi-step behavior. All deep-corpus rows carry per-step ground-truth engine states (15,791/15,791 steps nonzero) and position annotations of the content integers of reference proofs (100\% coverage); the latter serve the error analysis of Section VI. Isolation between corpus and test suites uses a two-level check: identifier level (exact matching of problem names and statement text, required to yield zero hits) and semantic level (an identity check after algebraic normalization, plus a statement-text four-gram overlap check; rows with overlap of 0.9 or above are removed from the training set). This procedure once detected content-level overlaps when several corpus versions were merged and removed them all; the final training set and test sets share zero problems.
\subsection{Test Suite}
The deep-chain test suite is produced from the generator candidate pool by a triple screen: unsolvable by the rule-based solver working from the statement text (so this pattern-matching baseline cannot answer it), in possession of a reference proof, and passing the overlap check against the training corpus. Of 338 candidates, 218 survive overlap filtering: four overlap counts of increasing stringency flag 7, 120, 217, and 326 candidates, and the adopted screen removes the 120 problems flagged at the second count, keeping a wider contamination margin than the loosest count; the survivors are 13 nested-induction problems, 24 lemma-tower problems, and 181 sum-of-squares problems. A shallow-chain suite (1,083 problems, chains of one to three steps) serves to evaluate early configurations and as the baseline reference of this paper. Both zero-capability controls score zero on the shallow suite: the untuned backbone and the rule-based solver each have 1,080 applicable problems and fail all of them (3 inapplicable problems excluded), showing that certified scores cannot be obtained by guessing or pattern matching.
\subsection{Configuration Overview}
The main configurations reported in this paper (Experiments 1 to 4) share the same backbone, training hyperparameters, batch order, and random seed; they differ only in the variables listed in Table VI: how proof text is produced, the chain-end loss weight, and the state timing of the decision frame. Experiments 1 to 3 isolate one factor at a time: Experiment 1 examines direct generation; Experiment 2 adds structured numeric output on top of it; Experiment 3 switches proof text to the compiler. Experiment 4 is a $2\times2$ factorial design. Experiment 5 is the external-model comparison.
\subsection{Review Validity}
Proportional readings reported with an interval use the Wilson 95\% confidence interval (CI) \cite{wilson}, where n is the number of graded problems and $z=1.96$:
\begin{table*}[!t]
\caption{Main experimental configurations. Experiments 1 to 4 share the backbone (SmolLM2-135M), training hyperparameters, batch order, and random seed; the four configurations of Experiment 4 differ only in the listed variables. Experiment 5 is an external comparison model and is not subject to these controls. Exp. denotes experiment.}
\centering
\scriptsize
\setlength{\tabcolsep}{3.5pt}
\begin{tabular}{>{\centering\arraybackslash}p{0.06\textwidth}>{\centering\arraybackslash}p{0.22\textwidth}>{\centering\arraybackslash}p{0.09\textwidth}>{\centering\arraybackslash}p{0.20\textwidth}>{\centering\arraybackslash}p{0.18\textwidth}}
\toprule
\textit{ID} & \textit{Proof-text production} & \textit{Chain-end loss weight} & \textit{Decision-frame state} & \textit{Purpose} \\
\midrule
Exp. 1 & Model-generated & 1.0 & Original (lagged by one step) & Direct-generation baseline \\
Exp. 2 & Model-generated + value slotting & 1.0 & Original & Isolates the numeric-generation mode \\
Exp. 3 & Compiler-generated & 1.0 & Original & Isolates the text-production mode \\
Exp. 4A & Compiler-generated & 1.0 & Original & Factorial baseline (same as Exp. 3) \\
Exp. 4B & Compiler-generated & 5.0 & Original & Factorial: loss weighting \\
Exp. 4C & Compiler-generated & 5.0 & Shifted to current state & Factorial: weighting + state \\
Exp. 4D & Compiler-generated & 1.0 & Shifted to current state & Factorial: state (final configuration) \\
Exp. 5 & DeepSeek-Prover-V1.5-RL direct generation & N/A & N/A & External control \\
\bottomrule
\end{tabular}
\end{table*}
\begin{equation}\mathrm{CI}(\hat{p}) = \frac{\hat{p} + z^{2}/(2n) \pm z\,\sqrt{\hat{p}(1-\hat{p})/n + z^{2}/(4n^{2})}}{1 + z^{2}/n}.\end{equation}
The interval is solved from the pivot of a binomial proportion: by the central limit theorem, $\sqrt{n}\,(\hat{p}-p)/\sqrt{p(1-p)}$ converges in distribution to the standard normal, and the interval endpoints solve the inequality $|\hat{p}-p|\leq z\sqrt{p(1-p)/n}$. Squaring both sides and rearranging gives the quadratic inequality in p
\begin{equation*}(1 + z^{2}/n)\,p^{2} - 2\,(\hat{p} + z^{2}/(2n))\,p + \hat{p}^{2} \leq 0;\end{equation*}
taking the two roots by the quadratic formula, with $(\hat{p}+z^{2}/(2n))^{2}-(1+z^{2}/n)\hat{p}^{2}=z^{2}\hat{p}(1-\hat{p})/n+z^{4}/(4n^{2})$ under the radical, yields the endpoint expression of (13).
Beyond the dual-protocol recheck of Section III-E, the following discipline applies to all readings. Every controlled-experiment score states the number of graded problems and the total; unexecuted reviews are never counted as zero; stratified recheck samples below the planned size are marked as such; the public-library readings of Section X report rates by arm, with the per-row ledgers released. Wherever rendering accuracy, the per-token agreement of the produced proof text with the reference proof, serves as the metric, a zero-parameter mode baseline (outputting the most frequent answer in the training set) is reported alongside: on the shallow-chain suite the final shallow-chain configuration reaches 0.9755 against 0.6276 for the baseline, a net gain of 34.8 percentage points, which rules out inflation from the most frequent answer. The zero scores of the two zero-capability controls (untuned backbone and rule-based solver), together with reverse verification (all corrupted proofs rejected), show that the review program rejects wrong proofs and passes reference proofs.
\section{Baselines and Negative Results in the Shallow-Chain Era}
With the curriculum corpus in place, the shallow-chain suite (1,083 problems, chains of one to three steps) produced the first baseline: the final shallow-chain configuration certifies 180/1,083 (16.6\%), while the two zero-capability controls, the untuned backbone and the rule-based solver, each have 1,080 applicable problems and score zero, so certified scores cannot come from guessing or pattern matching. Rendering accuracy is 0.9755 against 0.6276 for the zero-parameter mode baseline, a net gain of 34.8 percentage points; inflation from the most frequent answer is excluded.
The first mechanism-level ablation was run in the same period: adding step-level attention to the chain planner (a first-generation implementation) cost a net 62 problems over the full suite (118/1,083 against 180/1,083, same suite, seed, and batch order). This negative result points in the same direction as the state-shift factorial of Section VIII: the bottleneck does not lie in the expressive power of the decision mechanism.
Shallow-chain tasks are solvable, but chains of one to three steps cannot examine multi-step behavior. The deep-chain corpus and the 218-problem deep-chain exam (Sections IV-B and IV-C) then moved the examination ground to a qualitative change; Section VI opens with the first deep-chain exam.
\section{The Error-Product Structure of Direct Long-Proof Generation}
The first deep-chain exam, run once the deep-chain corpus and the 218-problem suite were in place, produced this paper's central measurement. Experiment 1 evaluates direct generation: the model plans the solution chain step by step, and a text-generation head writes the complete proof; certified passes number 0/218. The zero does not come from failed chain decisions: the first-operator hit rate is 218/218 and the exact whole-chain hit rate is 117/218 (53.7\%), with the nested-induction family 13/13 and the lemma-tower family 24/24 all correct. Meanwhile, the model's token accuracy on 48 held-out rows from the training distribution is 0.9637, in sharp contrast to the zero pass rate. Decomposing the generated text identifies the cause of failure. Counting the content integers (coefficients, exponents, constants) of each proof individually: on the 20-problem held-out control set drawn from the same generators, the probability that one content integer is reproduced exactly is about 0.79, and each proof carries about 14.9 content integers on average. Writing the content integers of a proof as $X_1,\ldots,X_k$, with integer i reproduced correctly with independent probability p${_{i}}$, proof validity requires every integer to be correct:
\begin{equation}\mathrm{Pass} = \prod_{i=1}^{k} p_i.\end{equation}
When all integers share the accuracy p, the expression equals $p^{k}$; taking p as an upper bound on every integer gives the power-law bound:
\begin{equation}\mathrm{Pass} \leq p^{k}.\end{equation}
\emph{Proposition 1 (product law and power-law bound):} suppose a proof contains k content integers, integer i is written correctly with independent probability p${_{i}}$, and certification requires all integers to be correct; then the pass rate is given by (14), and if no integer exceeds accuracy p, the bound of (15) holds.
\emph{Proof:} the certification event is the intersection of the k events that integer i is correct; by independence, its probability is the product of the per-integer probabilities, which is (14). If every integer's accuracy is at most p, each factor of the product is at most p, so the product is at most $p^{k}$, which is (15). \hfill$\blacksquare$
\emph{Corollary (load correction):} let $S\subseteq\{1,\ldots,k\}$ be the set of load-bearing integers determined mechanically by the controlled-corruption experiment; changing the value of an integer outside S does not change the certification outcome. The product therefore only needs its factors in S, Pass = $\prod_{i\in S}p_i$; when all load-bearing integers share the accuracy p, this is (17).
Substituting $p=0.7919$ and $k=14.9$ gives a pass rate of about 0.031, that is, less than one expected pass in 20 problems; the measured one pass is consistent with the prediction (Fig. 3(a)); at one pass in 20 problems this reading is a spot check, and the controlled-corruption experiment below carries the direct test of the product structure. On the test suite, the per-integer accuracy of the sum-of-squares family is only 0.0911; the same computation puts the pass rate near zero, consistent with the measured 0/181. In other words, the validity of a long proof is the conjunction of its load-bearing content elements, and when per-integer accuracy is insufficient the pass rate falls exponentially with the element count; token-level metrics do not reflect this structure and therefore do not constitute a valid performance measure in this setting. Enlarging the thinking budget does not help the structure either: in an auxiliary budget comparison on a synthetic task, doubling the multi-step budget left accuracy flat across levels (about 0.974), consistent with the ceiling lying in the conjunctive structure of the output channel rather than in the budget.
Law 1, the product structure, admits a direct test that does not depend on the model. For the annotated content-integer positions in the reference-proof pool (218 test problems plus 2,399 corpus rows, 2,617 proofs in total), each integer is independently replaced, with probability $1-p$, by a random value of the same magnitude (excluding the original value; the sign is kept), and the corrupted proofs are graded by the same review program. The design crosses six values of p from 0.99 to 0.50 with bins of the per-problem content-integer count (bin means 10.4, 15.1, 20.0, and 30.0), for 3,570 sampling runs and 1,680 corrupted units graded (0 review anomalies). A damaged unit survives if its host proof still certifies; Fig. 3(b) plots the per-bin survival of damaged units as a function of the corruption-design accuracy p, with the independence-theory curve 1${- p}$ overlaid; the pooled whole-proof rates are given in the caption. The theoretical mean weights the actual sampling structure: j runs over the sampled runs, each drawing its accuracy value p from the six design values and its integer count $k_j$ from the pool:
\begin{equation}T(p) = \frac{1}{N}\sum_{j=1}^{N} p^{k_j}.\end{equation}
The measured pass rate over the full pool is 0.558, slightly above the theoretical value of 0.530, a relative excess of about 5\% (Fig. 3(b)); both rates pool the 3,570 whole-proof draws, while the per-bin points use the 1,680 graded damaged units.
Slot-level tallies locate the source of the deviation: no corrupted unit at coefficient or exponent positions survives (0/432), and all 1,164 corrupted units of the nested-induction and lemma-tower families are rejected by review; every instance of tolerance occurs at multiplier and linear-form constant positions of the sum-of-squares family. In that family's proofs the multiplier appears only as a positive scalar, which the nonlinear-arithmetic tactic can absorb, so replacing the number leaves the proof valid (25/25 single-slot corruptions survive; tolerance occurs only at such positions of this family and is a property of the proof template). Accordingly, write $S\subseteq\{1,\ldots,k\}$ for the set of load-bearing integers, determined by the proof template and measured mechanically by the controlled-corruption experiment ($|S|\leq k$); a more precise statement of the pass rate is:
\begin{equation}\mathrm{Pass} \approx p^{|S|}.\end{equation}
The controlled-corruption experiment also provides a mechanical way to measure the load-bearing status of each integer position.
On this baseline, Experiment 2 changes the numeric output to structured slots: the content-integer positions of the proof template are predicted by the model in a discrete value space instead of being generated freely by the text head, and in one configuration the slot ground truth is supplied by the execution engine. The results: slots with engine supply pass 1/218, slots filled by the model pass 0/218, and the original baseline 0/218. After slotting, the model's fill-in accuracy reaches 94.5\% (39.6\% of rows fully correct), yet whole-problem certification stays near zero; the gap between high fill-in accuracy and failure under autoregression comes from the cumulative effect of earlier errors on later conditions in autoregressive generation. The result shows that local improvements in numeric generation (raising per-integer accuracy) cannot offset the exponential loss of the conjunctive structure; the free-generation stage itself must be removed.
\section{Compiler-Generated Proof Text}
The contrast between 53.7\% chain navigation and 0/218 certified direct generation pinpoints the text-generation stage. Experiment 3 changes the production of proof text to the compiler (Section III-D); the model outputs only the operator chain, and the text loss leaves the training objective. On the same suite, the certified pass rate rises from 1/218 to 117/218 (53.7\%). Moreover, the chains of the existing model (produced by the Experiment 2 configuration), when compiled, certify 121/218, matching the exact hit rate of those chains; the retrained pure-planning model gives 117/218, within four problems of the 121/218 obtained from the existing chains. These readings show that once free-text generation is removed, the certified pass rate of the system is determined entirely by chain-decision quality and the text stage no longer costs anything; the same chain set under the two text-production modes yields 1/218 versus 121/218, which directly quantifies the cost of the free-generation stage.
\begin{figure*}[!t]
\centering
\includegraphics[width=\textwidth]{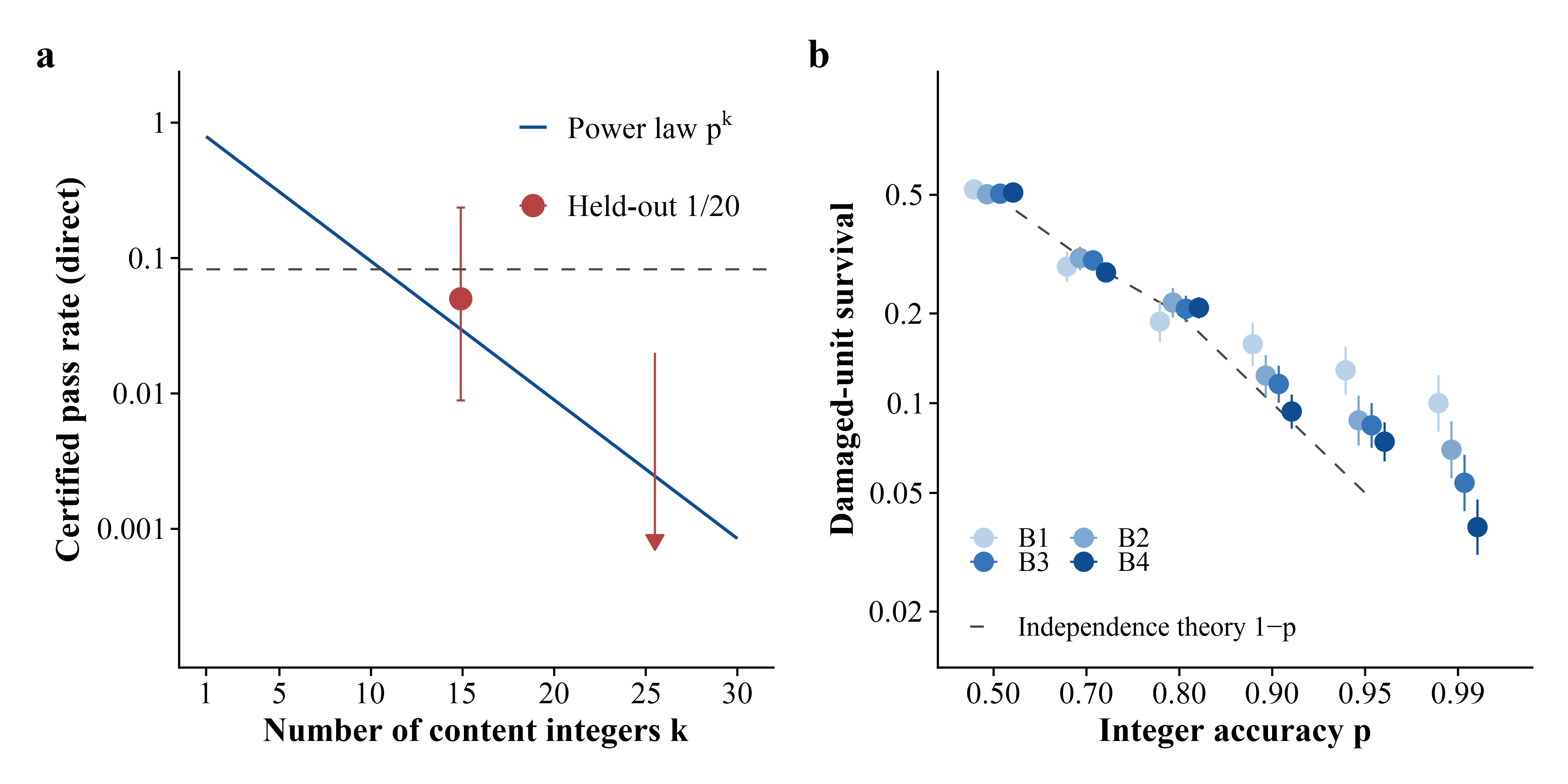}
\caption{Quantitative tests of Law 1. (a) The pass rate of direct generation decays as a power law in the number k of content integers (p = 0.79); the held-out control point (1/20, with a binomial 95\% interval) is overlaid, the dashed line marks the comparison model (18/218 = 8.3\%), and the red arrow marks the deep-chain suite at k = 25 (0/218). (b) Per-bin survival of damaged units (B1 to B4 by content-integer count) as a function of the corruption-design accuracy p, with the independence-theory curve 1${- p}$ overlaid; pooled over 3,570 whole-proof draws, the measured pass rate is 0.558 against the pooled theory 0.530.}
\end{figure*}
Given that compilation is a deterministic map, verified twice over (byte-identical reproduction on 2,399 corpus rows and 218/218 on reference chains), the certified pass rate of the system equals the correctness rate of the chains the model produces:
\begin{equation}\mathrm{Pass} = P(\text{chain correct}),\end{equation}
where chain correctness is measured by exact whole-chain match. The equality removes the text stage from the performance variables and leaves chain-decision quality as the only performance variable of the system: the 117/218 of Experiment 3 equals the exact chain hit rate, and the 121/218 of the Experiment 2 chains under compilation matches their hit rate as well. The product structure of Section VI is precisely the form that the right-hand side of this equality takes under the free-text paradigm.
\section{Attribution: A Factorial Analysis of Decision-Time State Information}
Compilation exposes all remaining errors at the final step of the chain. Localizing the 101 errors of Experiment 3 problem by problem and step by step: all 101 first errors occur at the last step of the chain, with no mid-chain errors, no premature termination, and no budget exhaustion; 55 of the erroneous steps carry predicted probability 0.9 or above, that is, the model is highly confident in its erroneous decisions. Combined with (5), the final class is fully determined by the residual value; the residual needed for the verdict is already available after the execution of step $T{-}1$, whereas the original decision frame carries the value from before the previous step, a one-step lag. Experiment 4 tests this hypothesis with a $2\times2$ factorial design: factor one is whether the state frame replaces the lagged value with the current one (the state shift below); factor two is the cross-entropy loss weight at the chain's final step (1.0 or 5.0). The four configurations are otherwise identical (same batch order, same random seed). Table VII reports the results.
Mathematically, the state shift augments the conditioning of (8):
\begin{equation}(o_t,a_t) = f_{\theta}\bigl(s, h_1, \ldots, h_{t-1}, R_{\mathrm{out}}^{(t-1)}\bigr),\end{equation}
where $R_{\mathrm{out}}^{(t-1)}$ is the residual after the execution of step $t{-}1$ per (4), computed before the decision and placed into the state frame; for the final step, $R_{\mathrm{out}}^{(T-1)}$ is exactly the final residual needed for the verdict.
Separating final-step quality from prefix quality yields a two-factor decomposition of the pass rate.
\emph{Proposition 2 (final-step conditional decomposition):} let B be the problem set whose first $T{-}1$ steps are all correct, and C the set whose whole chain is correct; then
\begin{equation}\mathrm{Pass} = A_{T-1}\cdot q_{T},\end{equation}
where $A_{T-1}$ = P(B) is the prefix accuracy and $q_{T}$ = P(C | B) the conditional accuracy of the final step.
\emph{Proof:} a fully correct chain implies a correct prefix, that is, $C\subseteq B$, so $P(C)=P(B)\cdot P(C\mid B)$, which is the claim. \hfill$\blacksquare$
On Experiment 3 the decomposition takes concrete values: the 101 first errors all sit at step T, so the prefix accuracy is $(117+101)/218=1.000$ and the conditional final-step accuracy is $q_{T}$ = 117/218 = 0.537; after the state shift it is $191/218=0.876$. The difference between the two configurations, $218\times(0.876-0.537)=74$ problems, matches the measurement, so the gain comes from improved conditional quality at the final step; prefix quality is not a loss term in either configuration.
\begin{table}[H]
\caption{Factorial results (n = 218). The simple effect of the state shift is +74 problems; the simple effect of loss weighting is $-7$ (average main effects $+68$ and $-13$), and weighting still lowers the pass rate on top of the shift (191 to 172).}
\centering
\footnotesize
\setlength{\tabcolsep}{3pt}
\begin{tabular}{>{\centering\arraybackslash}p{0.19\columnwidth}>{\centering\arraybackslash}p{0.12\columnwidth}>{\centering\arraybackslash}p{0.12\columnwidth}>{\centering\arraybackslash}p{0.26\columnwidth}>{\centering\arraybackslash}p{0.13\columnwidth}}
\toprule
\textit{Configuration} & \textit{State shift} & \textit{Chain-end weight} & \textit{Certified passes (n = 218)} & \textit{$\Delta$ vs. baseline} \\
\midrule
Baseline (Exp. 3) & No & 1.0 & 117 & N/A \\
Weighted & No & 5.0 & 110 & $-7$ \\
Weighted + shift & Yes & 5.0 & 172 & +55 \\
Shift (final configuration) & Yes & 1.0 & 191 & +74 \\
\bottomrule
\end{tabular}
\end{table}
These results state the second empirical law: inside a restricted decision space, information visibility is the first-order determinant of correctness, and loss allocation does not account for the difference. Correcting the state timing alone lifts the pass rate from 53.7\% to 87.6\%, whereas up-weighting the loss at the site where errors concentrate is negative under both states (117 to 110 without the shift; 191 to 172 with it) and, in the unshifted weighted configuration, raises confidently wrong decisions from 55 in the baseline to 69 (of its 108 errors); in the final configuration the same count falls from 55 to 12 (Fig. 4(b)). Weighting adds no decision-relevant information; under weighting the confidence distribution of the errors is nearly unchanged while their number grows. The migration of error composition tracks the completeness of state information (Figs. 4(a) and 4(b)): the final configuration's 27 errors all sit at the final step and all are confusions between the square class and the other class (zero-residual verdicts are all correct, 31/31; zero-class-related confusions fall from 36 to 0; confidently wrong decisions fall from 55 to 12). The zero-residual case can be read directly off the residual value, while the square verdict requires the model to look up one of six squares, a representation-learning problem under information already in hand rather than missing information. The classification of (5) is itself mechanically decidable from the residual value; the final configuration keeps this verdict inside the model's decision space, so the certified score measures the planning chain end to end, and the 27 remaining errors sit exactly at this verdict. Per-step decision entropy quantifies this structure: early-chain decisions are near-deterministic (entropy below 0.001 nats) and real uncertainty concentrates in steps 4 to 6 (Fig. A6); failing problems have seven times the entropy of passing problems at the first divergent step (0.191 versus 0.027 nats), while the top-1 probability at that step still reaches 0.918 (Fig. A6), so measurable extra uncertainty and high-confidence error hold at once. Two further pieces of parameter-side evidence appear in the appendix: the parameter geometry of the operator head separates backbone operators from added operators rather than clustering by problem family (Fig. A3); the two-dimensional loss cross-section at the training convergence point sits on a wide, flat stretch of the basin floor, with corner loss 15.2 times the convergence-point loss (Fig. A4); complete versions of the first-error probability and the along-chain entropy appear in Figs. A5 and A6.
\section{Cross-Model Examination and Scale Decomposition}
Once the two laws closed on the same exam, two questions followed: is the collapse specific to small models, and what are the separate contributions of the architectural change and of model scale? Experiment 5 evaluates DeepSeek-Prover-V1.5-RL \cite{dsp15} (7B parameters, int8 quantization, greedy decoding, zero-shot prompting, single generation capped at 1,024 tokens; completion-style prompts in the model's own quick-start format, 113 to 288 prompt tokens per problem) on the same 218-problem suite. Certified passes: 18/218 (8.3\%); 177 outputs contain an unsolved goal (sorry), and the remaining 23 failures carry no unsolved goal. Both systems are graded by the same frozen review program on the same compiled Lean environment (Section III-E). By family: nested induction 0/13, lemma tower 0/24, and every pass belongs to the sum-of-squares family (18/181). The final configuration of this paper reaches 191/218 on the same suite (Fig. 5(a)). The distribution of decision steps per problem appears in Fig. A7 (mean 6.12, chains of 5 to 9 steps; the two synthetic families have rigid chain lengths, and in the final configuration every remaining failure sits in the sum-of-squares family, Section VIII). This comparison must be read with the interface difference in view: the comparison model has no execution engine or compiler and must generate the complete proof as free text, so its failure mode matches the direct-generation experiments of Section VI; the suite demands more correctly written content integers than the model can produce. We therefore read this comparison as evidence on the structure of the output channel, not as a ranking of the two models' reasoning ability. Table VIII gives the resource comparison.
\begin{figure*}[!t]
\centering
\includegraphics[width=\textwidth]{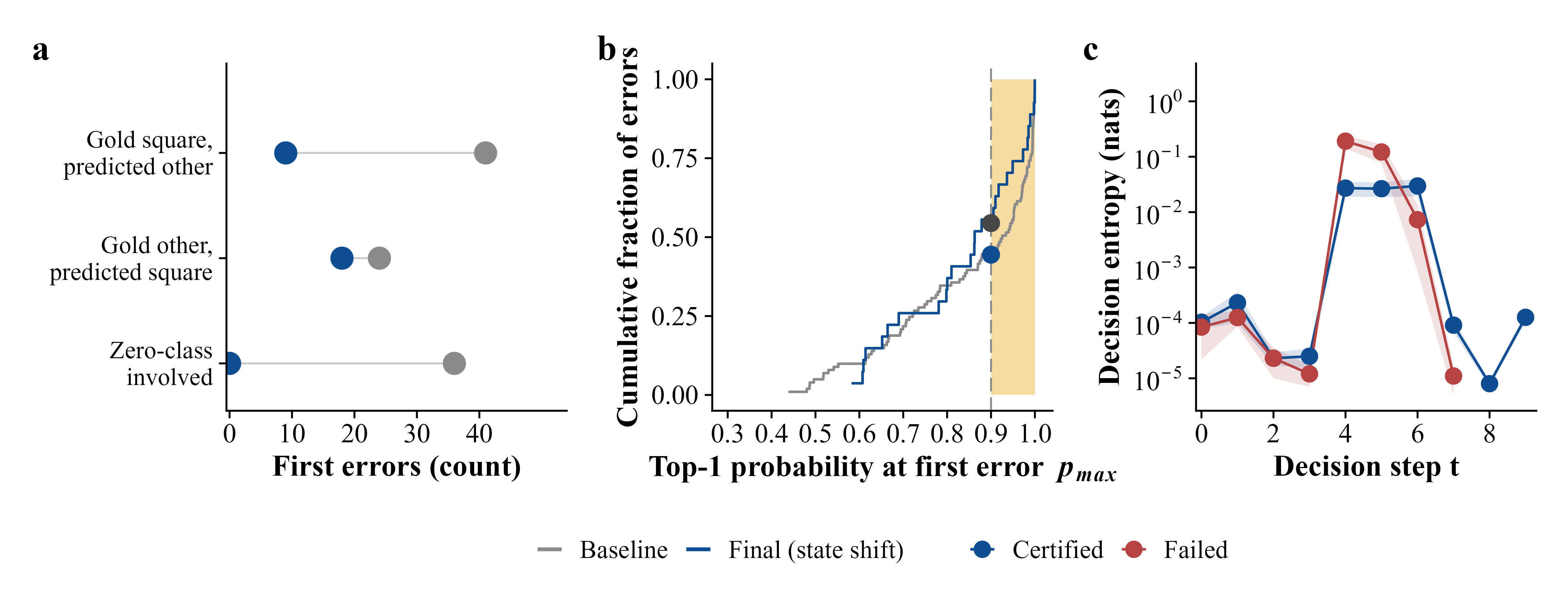}
\caption{Composition, confidence, and uncertainty of final-step errors (gray: baseline; blue: final configuration). (a) All 101 baseline first errors sit at the chain's final step; the per-class counts move from baseline to final configuration as 41 to 9, 24 to 18, and 36 to 0, and the 27 remaining errors of the final configuration are all table-lookup confusions between the square class and the other class, with all zero-residual verdicts correct (31/31). (b) Distributions of top-1 probability at the first erroneous decision step; the shaded band marks p ${\geq  0.9}$, and the two dots on the dashed line mark the confident-error fractions. The two configurations' shapes nearly coincide, and confidently wrong decisions fall from 55 to 12; the state shift changes the number of errors, not their confidence. (c) Decision entropy along the chain: early steps are near-deterministic (below 0.001 nats), real uncertainty concentrates in steps 4 to 6, and failing problems have seven times the entropy of passing problems at the first divergent step (0.191 versus 0.027 nats).}
\end{figure*}
A cross-generation recheck of the comparison uses a bidirectional dual suite: DeepSeek-Prover-V2-7B \cite{dspv2} (int4 quantization, greedy decoding, single generation capped at 8,192 tokens) answers both the main suite of this paper (218 problems) and a stratified suite of 500 rows drawn from the public library Lean-Workbook \cite{leanworkbook}, with problem sources and style close to that model family, so that a favorable skew of the exam cannot explain the comparison result. Both suites yield zero certified passes (0/218 and 0/500; Wilson 95\% upper bounds 0.017 and 0.008), and failing outputs almost all terminate in unsolved goals; generation-capacity triggers account for 77/218 and 177/500, and the problems that did not hit the cap (141 and 323) also all fail, so the zeros cannot be attributed to the generation budget. If the zeros simply reflected a domain mismatch, the model should do clearly better on the stratified suite close to its own problem sources; the measurement says otherwise. The preset failure decomposition (correlation of content-integer count with correctness) degenerates because both arms are constant at zero, and yields no readable quantity. This configuration generates 5,255 tokens per problem on average (3,919 of them in the thinking segment), takes 286 s, and peaks at 11,649 MiB; both generation volume and wall-clock are an order of magnitude above V1.5-RL. The readings of the two generations on the same main suite (18/218 and 0/218) with consistent interruption patterns show that the iteration did not change the fate of the direct-generation paradigm on this class of propositions. All verdicts came from the frozen review protocol of Section III-E.
Both ratios can be explained by the cost structures at the two ends. Per problem, the number of engine calls equals the number of decision steps T, and the number of model forward passes equals T up to one additional terminator decision; the text is compiled from templates and never enters the generation sampling loop, so the per-problem cost of this architecture is
\begin{equation}C_{\mathrm{sc}} = T\,(c_{\mathrm{fwd}} + c_{\mathrm{exec}}) + c_{\mathrm{compile}},\end{equation}
where $c_{\mathrm{fwd}}$ is the cost of one forward pass, $c_{\mathrm{exec}}$ the cost of one engine execution, and $c_{\mathrm{compile}}$ the cost of expanding the whole chain from templates; all three are deterministic and do not fluctuate with sampling. The direct-generation paradigm must generate every content integer digit by digit; its per-problem cost is
\begin{equation}C_{\mathrm{dir}} = N_{\mathrm{gen}}\,c_{\mathrm{tok}},\end{equation}
where $c_{\mathrm{tok}}$ is the cost of one generated token, and the measured mean of $N_{\mathrm{gen}}$ is 168.0 (V1.5-RL) and 5,254.6 (V2, including the thinking segment). The two cost expressions count different things: our corpus chains run 5 to 9 steps (corpus mean 6.58, exam mean 6.12), while the comparison models average 168.0 and 5,254.6 generated tokens; the measured ratios of Table VIII agree with this difference in counts (Fig. 5(b)).
\begin{figure*}[!t]
\centering
\includegraphics[width=\textwidth]{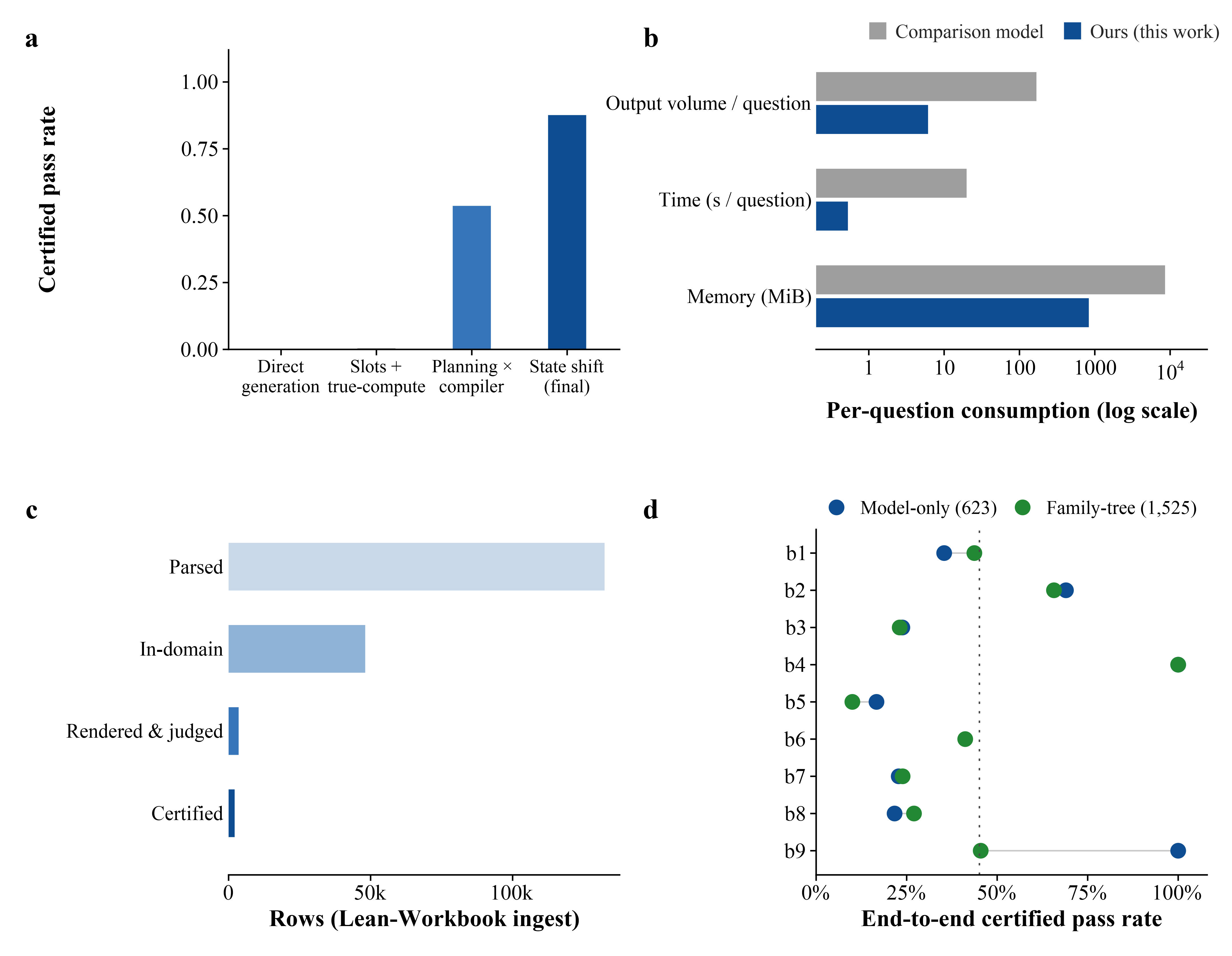}
\caption{Main results. (a) Certified pass-rate ladder of the main configurations: 0/218 (direct generation), 1/218 (slotting), 117/218 (compilation), 191/218 (state shift); for reference, the final shallow-chain configuration certifies 180/1,083 and the comparison model 18/218. (b) Resource comparison (log axis): output volume per problem (6.12 decision actions versus 168.0 generated tokens, 27.5${\times )}$, wall-clock per problem (0.53 s versus 19.88 s, 37.5${\times )}$, and peak memory (833 MiB versus 8,567 MiB, 10.3${\times )}$; the comparison model runs one to two orders of magnitude higher on all three. (c) The three-tier triage funnel over the full Lean-Workbook library ($132{,}378\to48{,}106\to3{,}536$, with 2,132 certified): the two removal steps are 84,267 logged out-of-domain rows and 5 rows that fail parsing in our ingest (7,448 library-corrupt rows are excluded before the 132,378 denominator); not executed does not mean scored zero, and the certified pass rate is 60.29\% (Wilson interval 58.67 to 61.89). (d) End-to-end comparison of model-only chain walking against family-tree chains over the nine extension families (45.10\% on the 623 model-renderable rows versus 44.07\% on the 1,525 family-tree rows).}
\end{figure*}
\begin{table*}[!t]
\caption{Resource comparison (n = 218). V1.5-RL and V2 denote DeepSeek-Prover-V1.5-RL and DeepSeek-Prover-V2-7B; the ratio column is each generation against this architecture; the V2 column is under int4 quantization with single generation capped at 8,192 tokens. This architecture produces no natural-language thinking text; 6.12 is the mean number of decision actions per problem (the terminator decision is additional).}
\centering
\scriptsize
\setlength{\tabcolsep}{3.5pt}
\begin{tabular}{>{\centering\arraybackslash}p{0.12\textwidth}>{\centering\arraybackslash}p{0.16\textwidth}>{\centering\arraybackslash}p{0.18\textwidth}>{\centering\arraybackslash}p{0.24\textwidth}>{\centering\arraybackslash}p{0.11\textwidth}}
\toprule
\textit{Metric} & \textit{This architecture (135M)} & \textit{V1.5-RL (7B)} & \textit{V2-7B (int4)} & \textit{Ratio (V1.5 / V2)} \\
\midrule
Certified passes & 191/218 (87.6\%) & 18/218 (8.3\%) & 0/218 & N/A \\
Output volume per problem & 6.12 (decision actions; no generated text) & 168.0 generated; 156.6 input tokens & 5,254.6 generated (incl. 3,918.8 thinking); 217.6 input tokens & $27.5\times$ / $858\times$ \\
Wall-clock per problem & 0.53 s & 19.88 s & 285.9 s & $37.5\times$ / $539\times$ \\
Peak inference memory & 833 MiB & 8,567 MiB & 11,649 MiB & $10.3\times$ / $14.0\times$ \\
\bottomrule
\end{tabular}
\end{table*}
As an injection control, the same step-by-step state information used by our system is appended to the external comparison model's prompt in three variants; the certified counts are 18, 14, and 13 out of 218, at or below the 18/218 obtained without injection. State text in the prompt does not substitute for the compiled decision channel. A three-configuration comparison gives a preliminary separation of the effect of scale from that of loadout: the inherited-head 135M, the new-head 1B, and the same-loadout new-head 135M reach 191, 170, and 180 out of 218; loadout migration and backbone scale each contribute about half the difference ($-11$ and $-10$ problems). Each configuration here is a single training run on the fixed suite; with n = 218, the binomial standard deviation of a pass count is at most about 7 (at most about 10 for the difference of two configurations), so we read the +74 state-shift effect as decisive and treat the smaller contrasts as directional readings; the direction of the weighting effect replicates across both states ($-7$ without the shift, ${- 19}$ with it, Table VII). Together with the attention ablation of the shallow-chain era (Section V), these results say: the collapse is not limited to small models, and neither a larger backbone nor a matched loadout reaches the state-shift configuration; information visibility is the decisive variable.
\section{Domain Extension: Validation on a Public Problem Library}
All experiments so far have run inside the proposition domain constructed by this paper and on the 218-problem main suite. To examine whether the recipe transfers and what the admissible domain actually covers, we extend the operator library and the corpus to the public library Lean-Workbook (139,826 problems) \cite{leanworkbook}. The extension follows the same discipline as the main suite: training data are synthesized by problem-family generalization, and every library problem is checked for leakage against problem name, statement text, and constant vector; the checking program itself is validated by injection experiments; the exams always use one deterministic renderer, and every problem is graded by the same frozen review program.
On Lean-Workbook, 7,448 rows shipped with the library cannot be parsed by the library's own toolchain, matching the exclusion count in the library's documentation; they do not enter the denominator, and the remaining 132,378 rows form the full denominator. Structural triage produces three tiers: 48,106 rows admissible under our operator system; 84,267 rows parseable but without a matching problem family; and 5 rows that fail parsing in our ingest (Fig. 5(c)). Within the admissible domain, the test fixture can render reference chains for 3,536 rows, which enter review, and 2,132 certify (60.29\%, Wilson 95\% interval 58.67 to 61.89); the reference arm of the existing operator families stands at 73.04\% (per-row counts in the released ledger), and the family-tree chains of the nine extension families at 44.07\% (672 of 1,525 rows). Rows outside the admissible domain are listed separately and do not enter the score. This coverage boundary is set by mechanical checks: which problems lie inside the capability domain, which do not, and why, are themselves outputs the system can produce.
Part of the chains behind these whole-library readings comes from fixture-rendered chains and the family trees. To isolate the model's own contribution, we run a chain-walking exam, with no fixture support, on the 1,525 family-tree rows of the nine extension families, none of which intersects the training corpus. The model decides autonomously throughout, including the terminator; on the 623 rows whose chains it can render, 281 certify (45.10\%, Wilson 41.24 to 49.03), against 44.07\% for the family-tree chains; the two arms have different denominators (623 versus 1,525), so they are compared at the rate level (Fig. 5(d)). Of the 623 rows, 331 model chains match the family-tree reference chains exactly, and the rendered text of these 331 rows matches the fixture rendering line for line by checksum (331/331), ruling out systematic drift in the rendering channel. The deep-chain exam reading of 191/218 (87.6\%) holds under the same-distribution condition: domain-extension training shows no statistically significant regression on existing capability. Against the re-measured starting model, the discordant pairs are $b=4$ (the re-measurement passes while the new configuration fails) and $c=9$ (the reverse), a net difference of c ${-  b = 5}$ problems in favor of the extended configuration; the exact binomial test
\begin{equation}p = \min\!\left[1,\; 2^{\,1-(b+c)}\sum_{i=0}^{\min(b,c)} \binom{b+c}{i}\right].\end{equation}
Substituting gives p = 0.267; the difference is not statistically significant.
The formula is constructed as follows: under the null hypothesis (equal pass rates), the directions of the $b+c$ discordant pairs are independent and equiprobable, so the null distribution of the smaller-side count $X=\min(b,c)$ is binomial $B(b+c,1/2)$; the two-sided exact probability is $2\,P(X\leq\min(b,c))$, capped at 1. With $b=4$ and $c=9$ there are 13 discordant pairs: $\sum_{i=0}^{4}\binom{13}{i}=1+13+78+286+715=1{,}093$, so p = $2\times1{,}093/2^{13}$ = 0.267, consistent with (23).
One negative result is reported in full. Merging the corpus of the existing routing families into training, so that routing problems enter the model's capability domain, fails the zero-regression gate in all three mixing ratios (176, 184, and 172 out of 218, all below 191), with degradation not monotone in the ratio; we register this as a sensitivity of corpus mixing to the training trajectory, and model-level inclusion of the routing domain is left to future methodological work. In the whole-library readings, this domain's score comes from the reference arm of the existing operator families and is labeled as such. All readings also passed an independent final audit: stratified re-grading by independent re-graders with zero discrepancies, and item-by-item reconciliation of discordant rows and denominator scope.
\section{Discussion}
The two empirical laws merge into one design principle: for the multi-step numeric derivation studied here, restrict the model's free-generation range and make the tool state needed for the verdict available at decision time. The first law says that when numeric content is generated freely, the pass rate decays exponentially with the length of the derivation, and improving per-integer accuracy cannot compensate; the second says that inside a restricted decision space, information visibility is the first-order determinant of correctness while loss allocation is not. The experimental results agree with the division-of-labor hypothesis of the introduction, and they point to one architecture: the model handles policy (choosing steps and arguments), the symbolic execution engine handles computation, and template compilation handles text.
The analysis of the error structure also carries independent evaluation significance. Token-level accuracy and the surface correctness of chain-of-thought intermediate steps can badly overestimate ability on long-derivation tasks; in settings where proof validity is a Boolean conjunction, per-integer accuracy and the element count are the quantities directly tied to the pass rate; the controlled-corruption experiment further shows that the load-bearing integer count is what matters: when non-load-bearing numbers are present (which depends on the proof template), the measured pass rate runs above the value computed from the surface count. This analysis also offers a way to read the failure samples of other direct-generation systems: the interruption patterns of the two comparison generations on this suite match our direct-generation experiments (177 sorry interruptions for V1.5-RL; near-total sorry interruptions for V2-7B on both suites) and are explained by the conjunctive requirement on content elements, without assuming insufficient model scale.
For tool-augmented methods, our results emphasize the timing and content of tool-state feedback, beyond the availability of tools. In our factorial experiment the tool, the execution engine, is available in every configuration; the only difference is the computation timing of the state inside the decision frame, and that single difference moves the pass rate by 74 problems. This suggests that in settings with step-by-step tool calls, the design of state feedback should be evaluated as an independent design variable.
The effect of scale is separated to a first approximation by the three-configuration comparison: loadout migration and backbone scale each contribute about half the difference (${- 11}$ and ${- 10}$ problems, Section IX; single-run contrasts of this size are read as directional), whereas within this domain the combined effect of removing free generation and completing state information (from near zero to 87.6\%) far exceeds the parameter gap between the two backbones in the comparison; this contrast in magnitude is the substance of our claim. Replication on larger backbones and transfer of the recipe to more problem families remain future work. The auditability of this architecture follows directly from separating responsibilities: the model's output is restricted to symbolic decisions, computation lives in a deterministic engine, and text comes from template compilation; each can be checked independently. This makes both the failures and the successes of the system explainable down to specific stages; our localization of the error structure (Section VI) and of state timing (Section VIII) both rest on this property. Extracting human-readable symbolic accounts from trained networks is an active direction \cite{tnnlstownsend}; here the symbolic account is native to the architecture. For applications that need reliability guarantees, this property has value independent of the pass rate; formal verification of neural networks is an established line in this journal \cite{tnnlsxiang}, \cite{tnnlsinn}, and auditability extends the same concern from the network to the process that generates its outputs. The structured process also changes how evaluation and scrutiny work. When the smallest unit of model behavior is an executable operator decision, a failure sample no longer offers only an aggregate signal but decomposes into position (which step failed), content (which argument or class failed), and confidence (the decision distribution); the error localization and attribution of Section VIII are built on this decomposition. For applications that presuppose reliability, this property matters as much as the pass rate itself: it allows mechanical enumeration and statistics of failure modes before deployment, without human sampling of generated text. Unlike chain-of-thought, this observability costs no natural-language output: the final configuration completes the entire solution within 6.12 discrete decision actions per problem on average, and the model generates no text to explain itself.
\section{Limitations}
This paper has several main limitations. The problem domain consists of constructively generated discrete-mathematics problems in three families (induction, inequalities, number theory); the conclusions do not transfer directly to open domains or to propositions requiring complex lemma rewriting. The main test suite has 218 problems and comes from our own generators; although zero-capability controls, semantic overlap filtering, and stating graded counts with every reading are in place, and the domain-extension validation on the public library (Section X) adds external readings, most rows of Lean-Workbook (84,267 without a matching problem family, plus the 5 ingest parse failures) lie outside the admissible domain of our operator system, and further extension of that domain is constrained by corpus-mixing sensitivity (the negative result at the end of Section X). Experiments use SmolLM2-135M as the main backbone; the three-configuration comparison involving the 1B backbone (Section IX) gives a preliminary split of loadout migration and scale at roughly half each, and systematic validation at larger scale is still pending. Review verdicts are bound to our frozen review program and Lean version; changing the proof assistant or the version requires redoing the dual-protocol recheck. Finally, the external models in the comparison experiments are not connected to our toolchain; the comparison speaks to the difference in output-channel structure and cannot be read as a same-interface comparison of the two models' ability.
\section{Conclusion}
This paper examined the failure mechanism of long-chain generation in language-model automated theorem proving inside a fully auditable discrete-mathematics domain, and proposed SCOPE, an architecture with the model as policy planner. The main results follow. Under direct generation of complete proofs, the pass rate is bounded by a power law in the accuracy of single content elements, and token-level accuracy diverges from the pass rate (token accuracy 0.9637 with 0/218 passing; the control-set pass rate is consistent with the power-law prediction, and the controlled-corruption experiment supports the constraint on 2,617 reference proofs and locates its deviations at non-load-bearing positions, with 0/432 coefficient- and exponent-position corruptions surviving). Replacing free-text generation with a compiler raises the pass rate from 1/218 (the slotting baseline) to 117/218, and the text stage stops costing anything. A factorial experiment on final-step decisions shows that the state timing of the decision frame alone determines a 74-problem difference in pass rate (117/218 versus 191/218), while loss weighting is negative. On the same suite, the 135M configuration with the toolchain attached reaches an 87.6\% certified pass rate at one to three orders of magnitude lower resource cost than two generations of 7B proof-specialized direct generators, and the cross-generation recheck points the same way (V2-7B certifies zero on both the main suite and the public-library stratified suite, including the problems that never hit the generation cap). All multi-step thinking happens in the discrete decision sequence, with no natural-language thinking text. We conclude that in the multi-step numeric derivation studied here, output-channel structure and decision-information visibility are more direct improvement variables than model scale. On the public library, domain extension reaches a 60.29\% certified pass rate on gradeable problems inside the admissible domain under the full pipeline, with no statistically significant regression on the old domain; on the extension families, the model's own chains certify at the level of the family-tree chains. Systematic validation on larger backbones is future work.
\appendices
\setcounter{figure}{0}
\renewcommand{\thefigure}{A\arabic{figure}}
\section{Figure and Table Inventory with Data Sources}
The five main figures are numbered in order of appearance. Fig. 1, the SCOPE architecture and data flow (division of labor among the three components, the trained versus parameter-free marking, the zero-token thinking label, and the main readings); Fig. 2, the low-dimensional projection of decision-frame hidden states (the measured family organization and chain progression); Fig. 3, quantitative tests of Law 1 (the power-law curve with measured points, and the controlled-corruption panel); Fig. 4, composition, confidence, and per-step entropy of final-step errors (three panels); Fig. 5, the main results grid (configuration pass-rate ladder, resource comparison, the Lean-Workbook triage funnel, and the nine-family end-to-end comparison). The carrier-comparison readings of multi-hop latent readout are carried by Table IV and the text of Section IV-A and are not given a separate figure. Seven appendix figures follow: Fig. A1, training dynamics of the final configuration (loss components and gradient norms, 6,000 steps); Fig. A2, the depth-curriculum ladder of the discretization experiment (gate-probe trajectory, Section IV-A); Fig. A3, the vocabulary geometry of the operator-selection head (separation of backbone and added operators, not family clustering); Fig. A4, the filtered-direction loss cross-section at the training convergence point ($51\times51$ grid, a wide basin around the convergence point); Fig. A5, complete two-configuration distributions of first-error probability (the full version of Fig. 4(b)); Fig. A6, the complete along-chain decision-entropy curves (the full version of Fig. 4(c)); Fig. A7, decision-step counts and per-family pass rates (Sections VII and IX). Per-figure data files for every figure (with source paths and checksums) live in the llm\_analysis directory of the code repository; every number can be exported directly from the raw files of the results and gen44 directories.
The eight tables are numbered in order of appearance: Table I, problem-domain layers and the operator vocabulary; Table II, execution logic of representative operators; Table III, step-by-step computation of a sum-of-squares problem; Table IV, carrier comparison for multi-hop latent readout; Table V, mechanism decomposition of the discretization experiment; Table VI, main experimental configurations; Table VII, factorial results; Table VIII, resource comparison. Data sources as above.
\begin{figure*}[!t]
\centering
\includegraphics[width=0.95\textwidth]{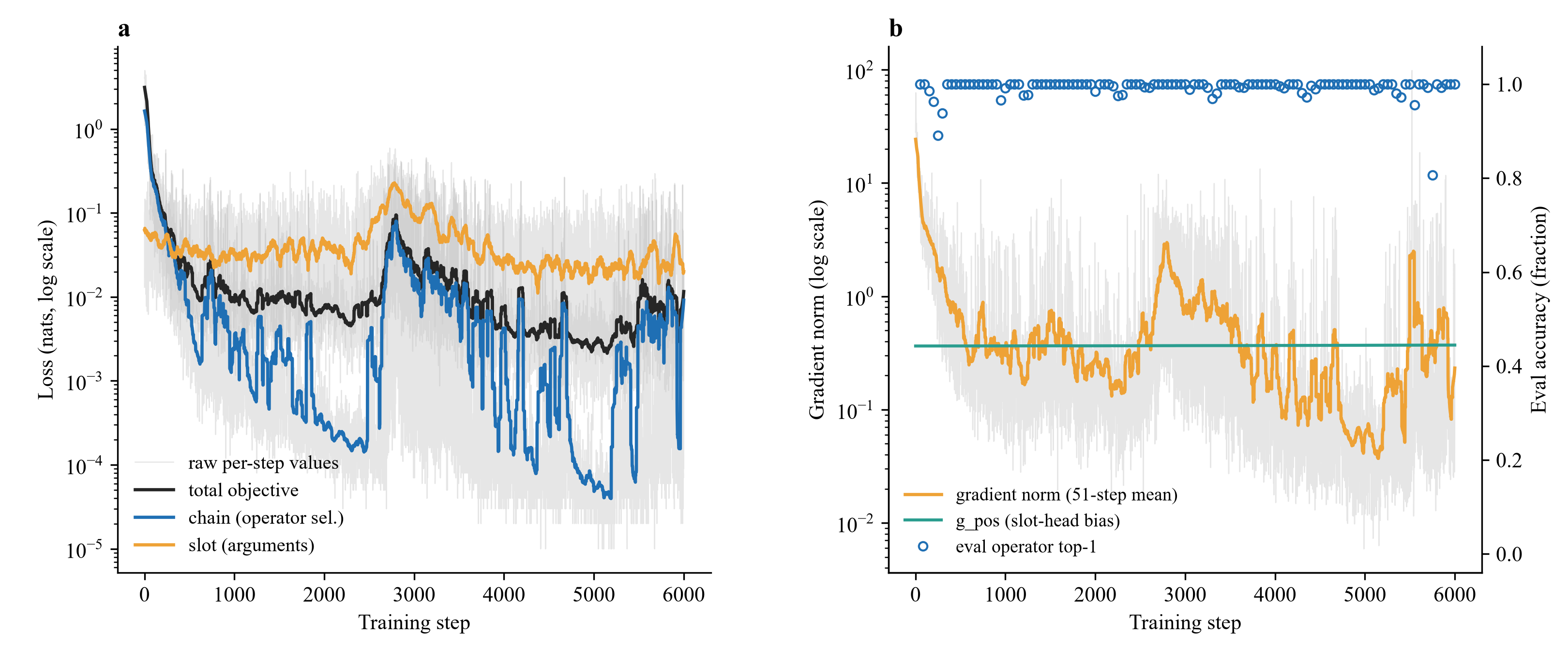}
\caption{Training dynamics of the final configuration (6,000 steps). (a) Total objective with the chain and slot components (log axis; gray: raw per-step values), falling from 4.03 to 0.025 nats. (b) Clipped gradient norm, the slot-head bias g\_pos (flat throughout), and held-out operator top-1 accuracy; the top-1 probe saturates at 1.00 from step 50 onward, so training progress is read from the loss. The text-modeling loss is not plotted (5.4 to 8.8 nats, different normalization and scale).}
\end{figure*}
\begin{figure*}[!t]
\centering
\includegraphics[width=0.95\textwidth]{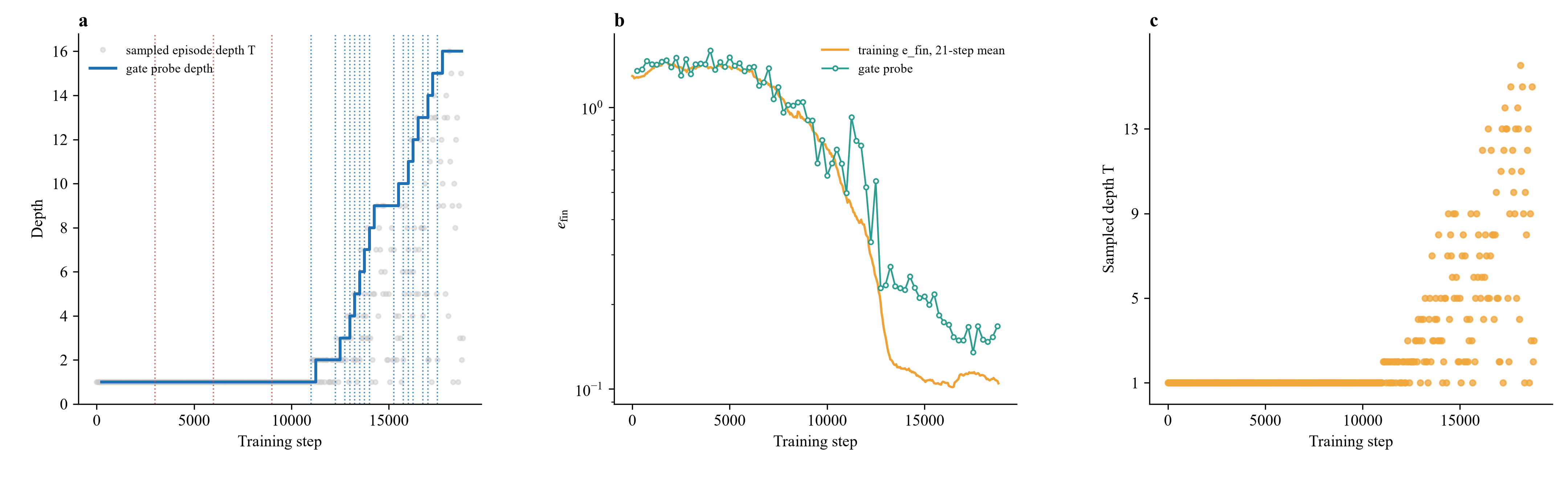}
\caption{The depth-curriculum ladder of the discretization experiment (v15). (a) The gate-probe measured depth rises monotonically from 1 to 16; the blue dotted lines mark the fifteen curriculum openings, all between training steps 11,000 and 17,500, the red dotted lines mark the three stall phases, and the gray points show depth sampling of training episodes. (b) Probe and training final-step error curves (log axis); the gate-probe error at depth 16 is 0.167, and the lower training reading (0.096) is a T = 3 training-step value, not the depth-16 result. (c) Depth sampling of training episodes; 241 of the 377 episodes remained sampled at $T=1$.}
\end{figure*}
\begin{figure*}[!t]
\centering
\includegraphics[width=\textwidth]{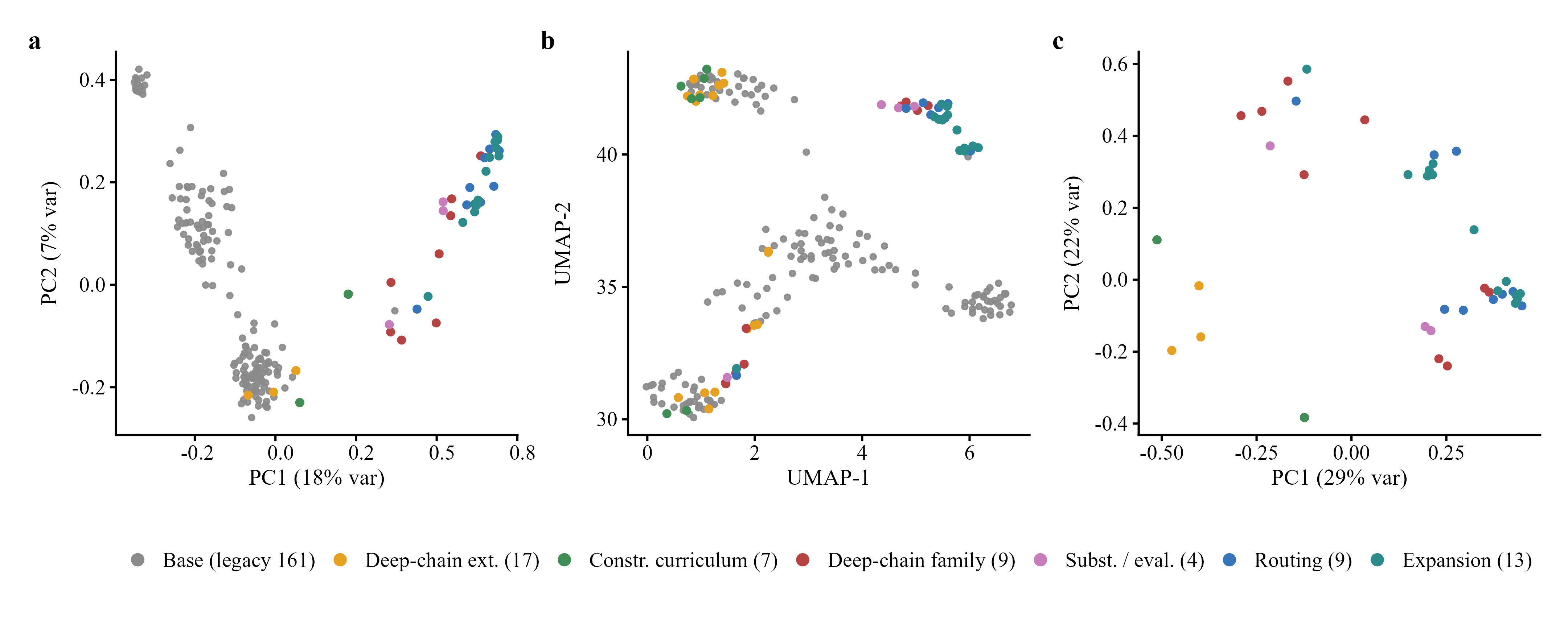}
\caption{Low-dimensional projection of the operator-selection head's output vectors (220 rows). (a) Principal component analysis (PCA) of all rows; (b) UMAP of all rows; (c) PCA of the 59 added rows. The head encodes the boundary between backbone and added operators, not a seven-segment clustering by problem family: the leave-one-out nearest-neighbor share is 0.969 within the backbone (156/161 rows), every other segment scores at most 0.462 (four segments score exactly 0.000), and within the 59 added rows the share is 0.407 against a 0.197 chance level; the seven-way silhouette is negative (${- 0.094}$ over all rows, ${- 0.083}$ over the added 59).}
\end{figure*}
\begin{figure*}[!t]
\centering
\includegraphics[width=\textwidth]{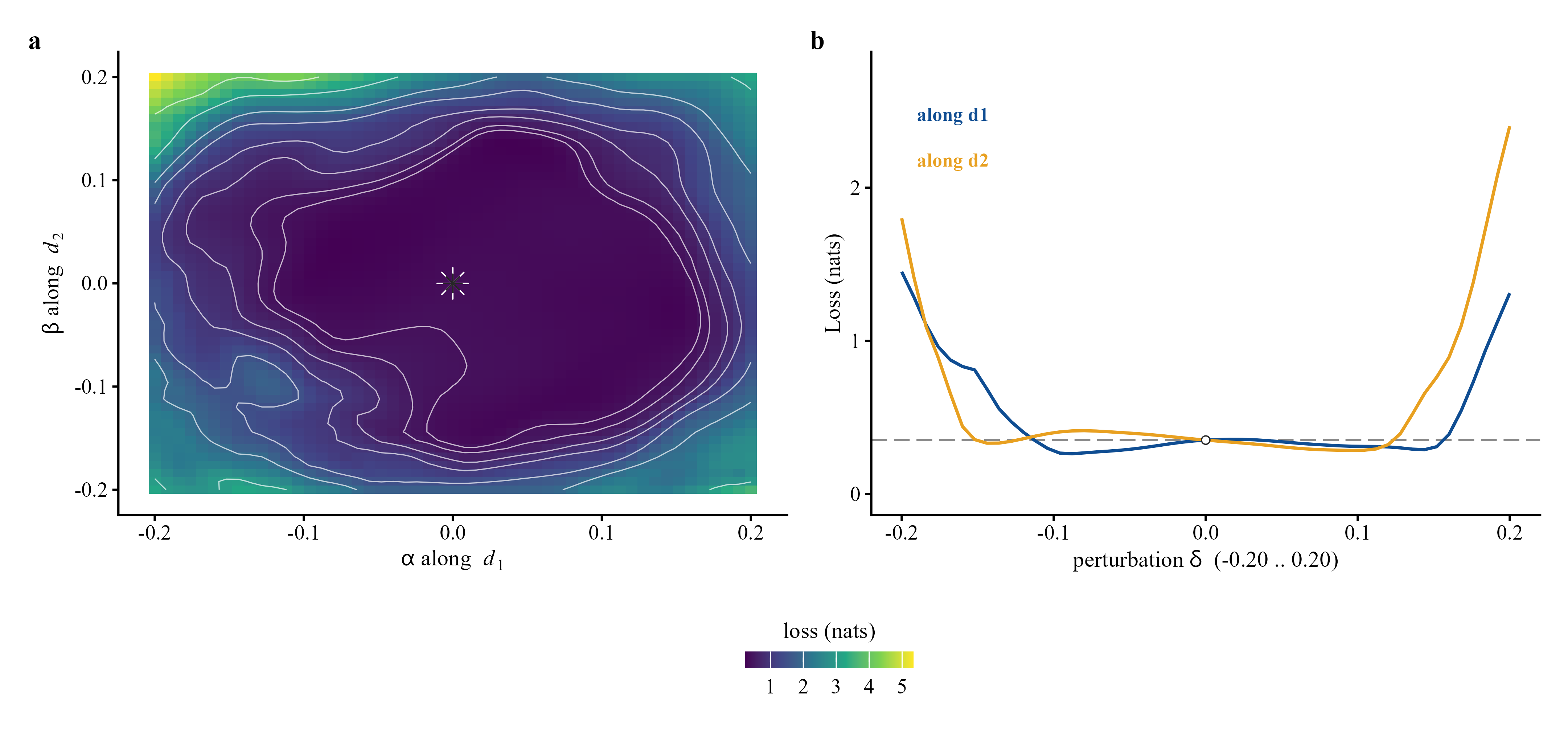}
\caption{Filtered-direction loss landscape at the training convergence point ($51\times51$ grid; filter-normalized following Li et al. \cite{liloss}; mean cross-entropy of the operator head at decision step k = 4 over 32 exam items with the reference prefix; the grid extends 0.20 along each filtered direction, an extent set by a one-dimensional calibration scan). (a) Filled contour map around the convergence point (star). (b) The two cross-sections along the filtered directions d1 and d2; the dashed line marks the convergence-point loss (0.351 nats). The section minimum (0.235 nats) lies below the convergence-point value, as expected when a two-dimensional filtered direction cuts through the full parameter space; the corners reach 5.34 nats, 15.2 times the convergence-point loss, and the loss stays within about 0.4 nats for perturbations up to about 0.1, so the convergence point sits on a wide, flat stretch of the basin floor.}
\end{figure*}
\begin{figure*}[!t]
\centering
\includegraphics[width=\textwidth]{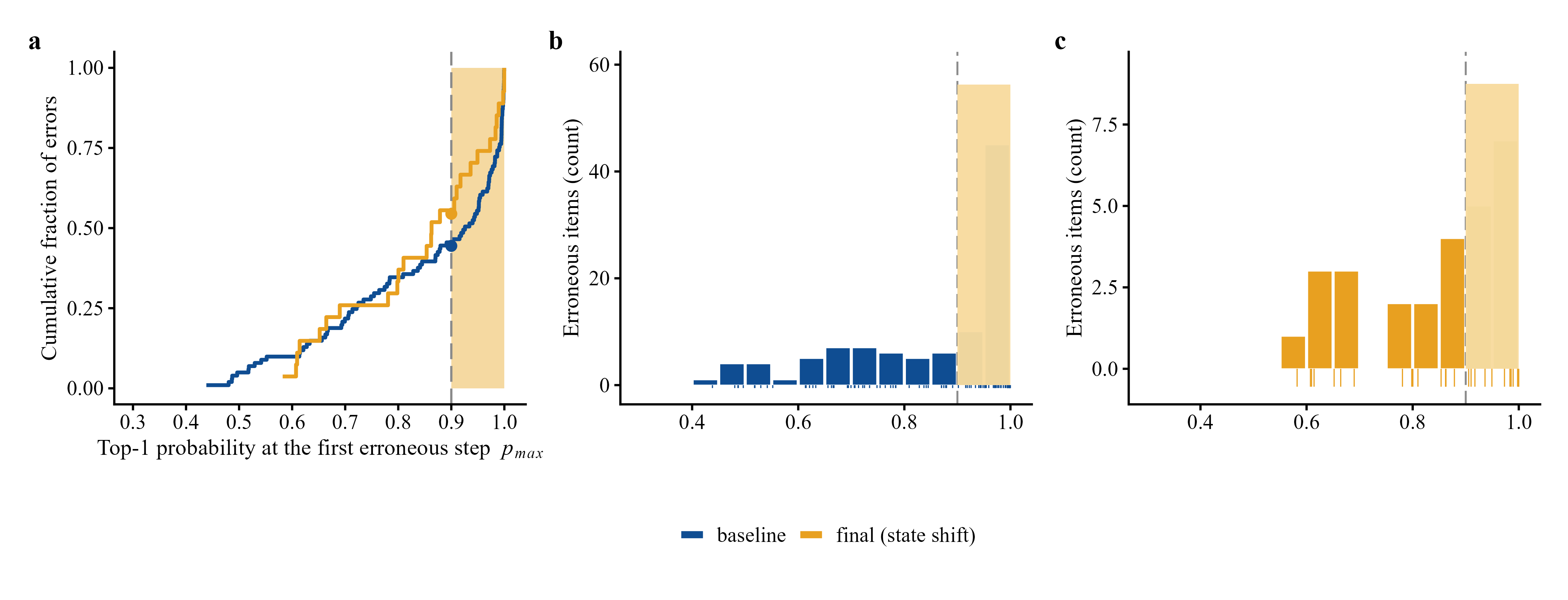}
\caption{Empirical distributions and counts of top-1 probability at the first erroneous decision step in the two configurations (baseline n = 101, final configuration n = 27; counts at the 0.9 threshold: 55 and 12); the full version of Fig. 4(b). (a) Cumulative distributions; the shaded band marks p ${\geq  0.9}$, and the dots on the dashed line mark the confident-error fractions. (b), (c) Absolute error counts per configuration in 0.05-wide bins, with rug marks showing the individual errors. The two configurations are paired in-series within the same evaluation pipeline (same source and grading rule as Table VII); a re-test reading (13/27) coexists and is not drawn.}
\end{figure*}
\begin{figure*}[!t]
\centering
\includegraphics[width=\textwidth]{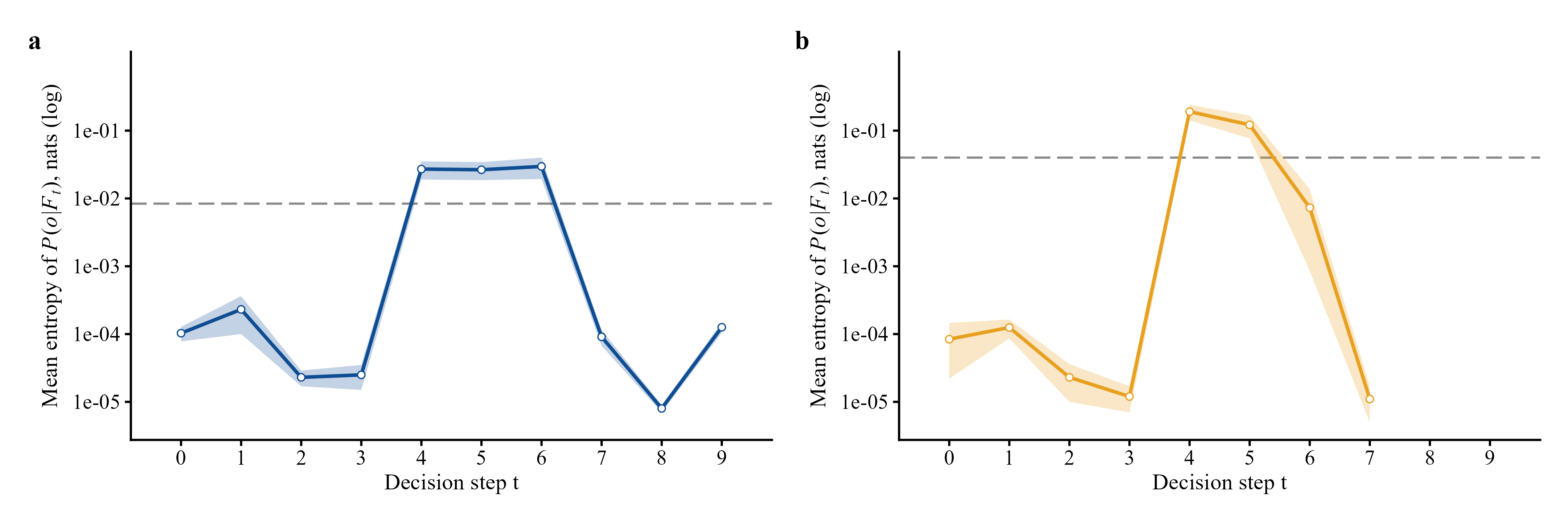}
\caption{Operator-decision entropy along the chain (log vertical axis), the full version of Fig. 4(c). (a) Certified items (191 items, 1,375 frames); (b) failed items (27 items, 177 frames); the dashed lines mark each group's overall mean (0.010 and 0.048 nats). At $t=4$ the two groups read $0.027$ and $0.191$ nats. The per-step frame counts are 123 at t = 6, 58 at t = 7, and 24 at t = 8 to 9 for the certified group and 11 at t = 6 and 4 at t = 7 for the failed group, so the failed group's late points rest on a few frames; the model is measurably less certain where it goes wrong, yet its top-1 probability there is still 0.918.}
\end{figure*}
\begin{figure*}[!t]
\centering
\includegraphics[width=\textwidth]{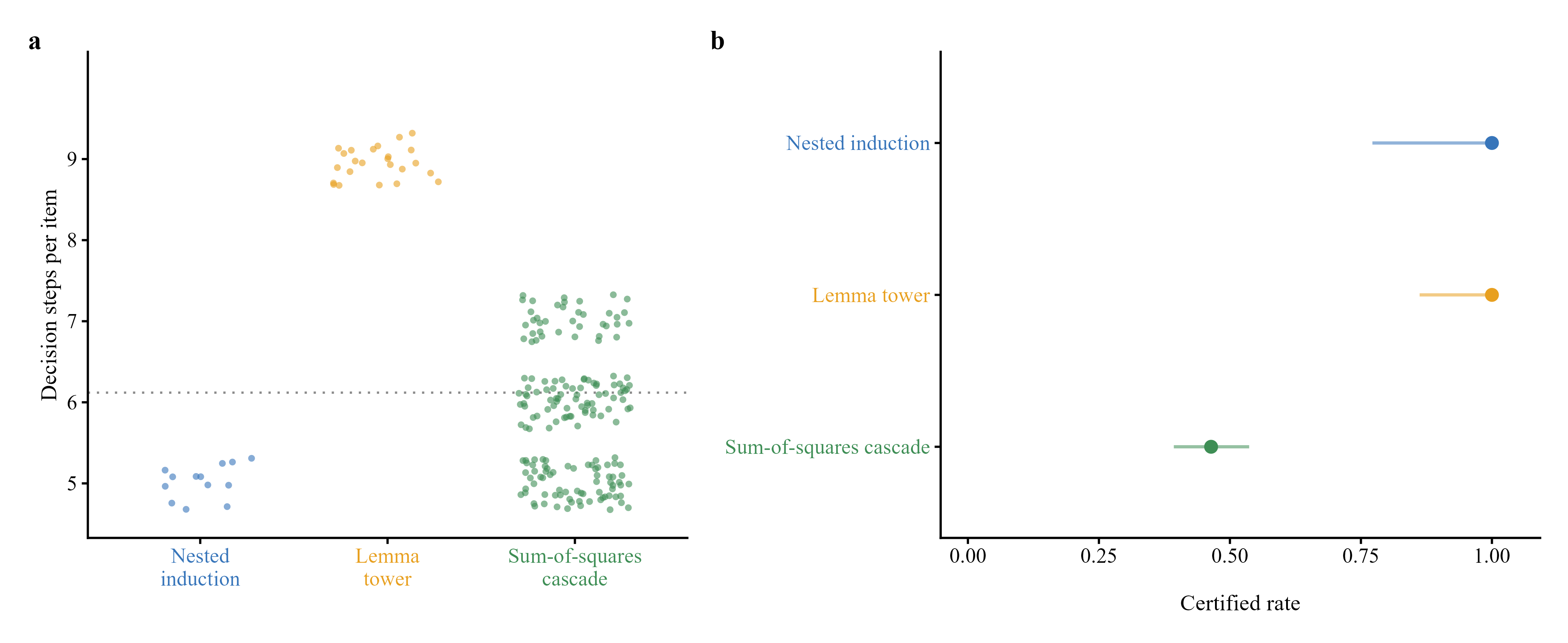}
\caption{Chain lengths and per-family certification of the 218 reference operator chains. (a) One dot per item at its decision-step count, jittered within the bin; the dotted line marks the pooled mean (6.12 steps); the sum-of-squares family spreads over 5 to 7 steps (71/72/38 items) while the other two families sit at the fixed lengths 5 and 9. (b) Certified pass rate of the model-produced chains per family with Wilson 95\% intervals computed from k/n (nested induction 13/13, interval 0.77 to 1.00; lemma tower 24/24, 0.86 to 1.00; sum-of-squares 84/181, 0.39 to 0.54), for the Experiment 2 chains under compilation (121/218); the reference-chain arm is 218/218 by construction.}
\end{figure*}
\section{Review Program and Key-File Checksums}
The review program source is drct\_common.py, version-frozen for the entire experimental period, MD5 7c31b58da9d7228cee31b9be581fe5dd. The raw reading files of every experiment (per-problem records, stratified rechecks, and reverse-verification results) and their checksums are listed in MANIFEST.md and the results directory of the code repository; training configurations and batch-order checksums are recorded in the report file of each experiment. The key files of the domain-extension and public-library validation (the final experiment report, the literal-scope full re-grading summary, the final-exam model checkpoint, and the final-audit report) and their checksums are in the gen44 directory; the key files of the cross-model dual-suite comparison (the experiment report, the grading summary, and the failure-decomposition artifacts) and their checksums are in the gen45 directory.
\clearpage
\clearpage
\IEEEtriggeratref{18}
\end{document}